\documentclass[10pt,twocolumn,letterpaper]{article}

\usepackage{iccv}              %

\usepackage{iccv}
\usepackage{times}
\usepackage{epsfig}
\usepackage{graphicx}
\usepackage{amsmath}
\usepackage{amssymb}
\usepackage{array}
\usepackage{multirow}
\usepackage{booktabs}
\usepackage{nicefrac}
\usepackage{subcaption}
\usepackage{siunitx}[2018/01/01]
\usepackage{colortbl}
 
\usepackage[accsupp]{axessibility}  %

\newcolumntype{Y}{>{\bfseries}S[table-format=2.1, mode=text, detect-weight=true, detect-inline-weight=true, detect-family=true]}

\def\ours{PseudoMapTrainer} %
\def\covratio{m}

\newcommand\scalemath[2]{\scalebox{#1}{\mbox{\ensuremath{\displaystyle #2}}}}

\definecolor{iccvblue}{rgb}{0.21,0.49,0.74}
\usepackage[pagebackref,breaklinks,colorlinks,allcolors=iccvblue]{hyperref}

\def\paperID{5303} %
\def\confName{ICCV}
\def\confYear{2025}

\title{\ours: Learning Online Mapping without HD Maps}

\author{Christian Löwens$^{1,3}$ \quad Thorben Funke$^{1}$ \quad  Jingchao Xie$^{1,4}$\quad Alexandru Paul Condurache$^{2,3}$\vspace{1mm}\\\normalsize
$^1$Bosch Research \quad $^2$Automated Driving, Bosch \quad $^3$University of Lübeck \quad $^4$Technical University of Munich\\
{\tt\small \{christian.loewens, thorben.funke\}@bosch.com}
}

\begin{document}
\maketitle
\begin{abstract}
Online mapping models show remarkable results in predicting vectorized maps from multi-view camera images only. However, all existing approaches still rely on ground-truth high-definition maps during training, which are expensive to obtain and often not geographically diverse enough for reliable generalization. In this work, we propose \ours, a novel approach to online mapping that uses pseudo-labels generated from unlabeled sensor data. We derive those pseudo-labels by reconstructing the road surface from multi-camera imagery using Gaussian splatting and semantics of a pre-trained 2D segmentation network. In addition, we introduce a mask-aware assignment algorithm and loss function to handle partially masked pseudo-labels, allowing for the first time the training of online mapping models without any ground-truth maps. Furthermore, our pseudo-labels can be effectively used to pre-train an online model in a semi-supervised manner to leverage large-scale unlabeled crowdsourced data. The code is available at \href{https://github.com/boschresearch/PseudoMapTrainer}{github.com/boschresearch/PseudoMapTrainer}.
\end{abstract}

\section{Introduction}
\label{sec:intro}
\begin{figure}[ht]
    \centering
    \includegraphics[clip, trim=15.8cm 10cm 14.5cm 1.5cm, width=\linewidth]{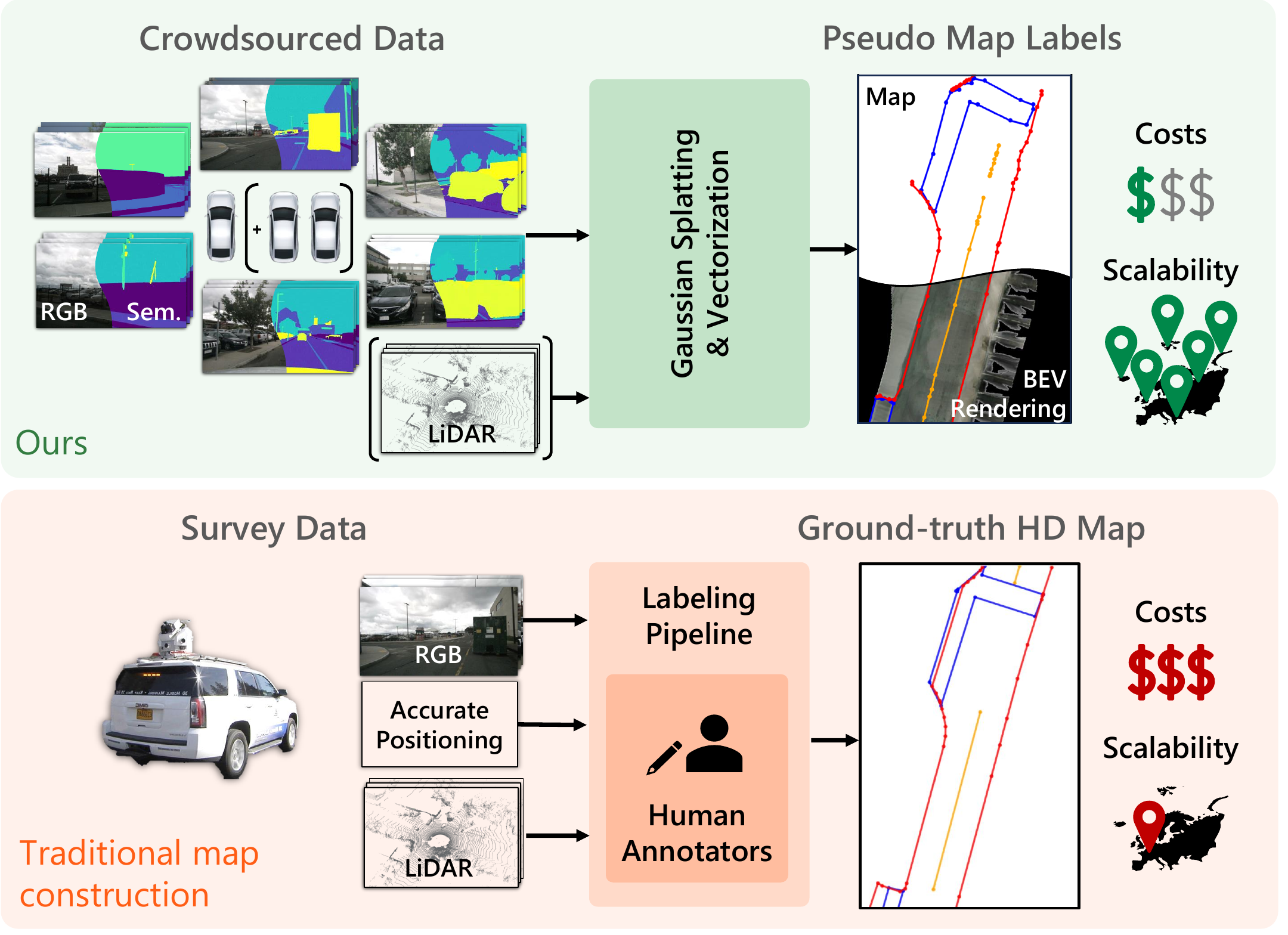}
    \caption{\textbf{Motivation for pseudo-labels.} Compared to conventional HD mapping, our method produces maps without human annotations via Gaussian splatting of surrounding camera and segmentation images, greatly reducing costs and enhancing scalability. We then use those labels to train an online mapping model. $[~\cdot~]$ denotes optional components. Survey vehicle from \cite{oregon2016survey}.}
    \label{fig:motivation}
\end{figure}

High-definition (HD) maps play a crucial role in autonomous driving, offering precise representations of road geometries, traffic signs, and other essential infrastructure \cite{ijgi13070232, Liu_Wang_Zhang_2020}. Traditionally, these maps are constructed from survey vehicles equipped with high-precision sensors and curated by human annotators. While accurate, this process is expensive and faces challenges in maintaining up-to-date information due to the dynamic nature of real-world environments \cite{ijgi13070232}.
To mitigate these limitations, the research community has increasingly focused on online mapping methods, which learn to generate maps in real time using only data from vehicle-mounted sensors \cite{li2022hdmapnet, liu2023vectormapnet, liao2022maptr,li2023lanesegnet}. A substantial challenge in this domain is the reliance on extensive ground-truth map labels for supervised learning, which are labor-intensive to produce (see Fig.~\ref{fig:motivation}) and often not geographically diverse enough for reliable generalization \cite{lilja2024localization}.

\begin{figure*}
    \centering
    \includegraphics[clip, trim=0.4cm 4.6cm 0.2cm 4.6cm, width=\textwidth]{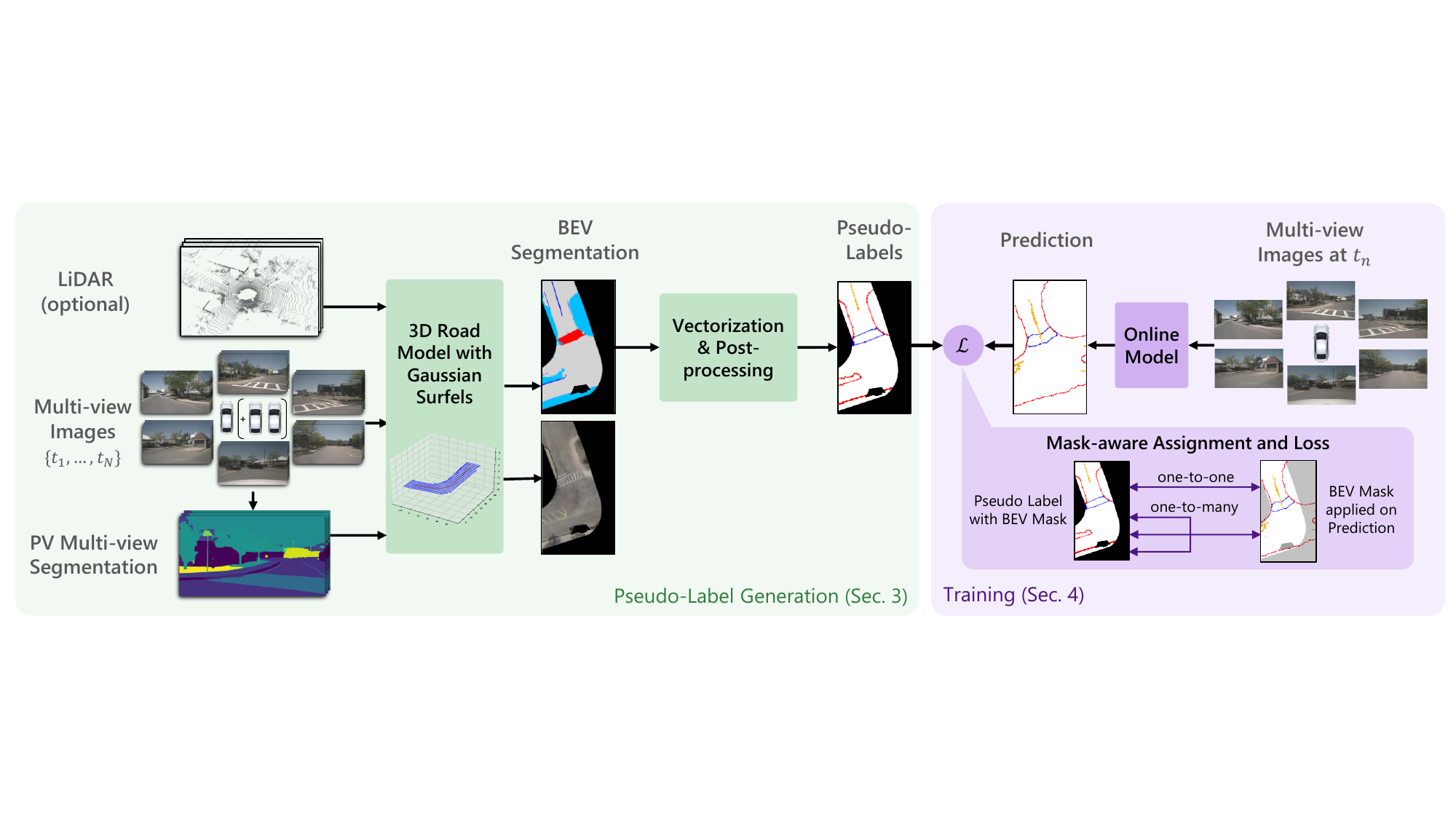}
    \caption{\textbf{Label generation and training pipeline of \ours.} We utilize sensor data from single or multiple trips and infer the corresponding 2D perspective view (PV) segmentations by a pre-trained network to build a coherent meshgrid of Gaussian surfels. Then, we render a BEV segmentation and extract the vectors as our pseudo-labels. Since these labels do not cover the full BEV range (see black regions), we train an online mapping model with a mask-aware one-to-many assignment algorithm and loss function.}
    \label{fig:pipeline}
\end{figure*}

Given that it is much easier to collect large amounts of unlabeled crowdsourced data from vehicles already on the road, the question arises whether this data can also be leveraged to train online mapping models. Therefore, in this work, we propose \ours, a novel approach that eliminates the need for ground-truth HD maps even during training. Specifically, we generate pseudo-labels based on road surface reconstruction using Gaussian splatting \cite{kerbl20233d}. We model the road surface as a mesh of Gaussian surfels~\cite{feng2024rogs} and optimize it with temporal multi-view camera images. Since surfels can encode not only geometric and color properties but also semantics, we can directly render semantic bird's-eye view (BEV) maps and subsequently derive vectorized pseudo map labels (see Fig.~\ref{fig:pipeline}).

As our pseudo-labels are derived from observations collected from a single or few vehicle passes, they are inherently incomplete. Thus, a key challenge when training with pseudo-labels is handling occlusions and missing data. To address this, we introduce a mask-aware assignment strategy and loss function enabling robust learning from incomplete labels. Furthermore, we explore the utility of pseudo-labels for semi-supervised learning. By pre-training our model on large-scale pseudo-labeled data and fine-tuning it on a limited set of ground-truth labels, we demonstrate significant performance improvements.

To the best of our knowledge, we propose the first approach to vectorized online mapping without any ground-truth HD map. Our contributions are:
\begin{itemize}
    \item A pseudo-label generation pipeline based on a road surface reconstruction using Gaussian surfels.
    \item A mask-aware assignment and loss function that robustly trains online models despite partial BEV observations.
    \item Improvement of online models with limited access to ground-truth labels via semi-supervised learning.
\end{itemize}

\section{Related Work}

\noindent\textbf{Road surface reconstruction.}
3D reconstruction methods based on structure-from-motion and multi-view stereo have shown strong performance in well-textured environments \cite{schoenberger2016mvs, schoenberger2016sfm}. However, these techniques often struggle with flat, low-texture road surfaces. Drawing inspiration from neural radiance fields (NeRF) \cite{mildenhall2020nerf}, recent approaches use an explicit mesh to model the road geometry, while elevation \cite{mei2024rome} and color \cite{wu2024emie} are represented implicitly. Since these methods exhibit a high computational demand, 3D Gaussian splatting \cite{kerbl20233d} has emerged as an efficient alternative representing scenes using Gaussian spheres. To reconstruct surfaces, these spheres can be reduced to flat surfels~\cite{huang20242d,dai2024high}. Accordingly, RoGs \cite{feng2024rogs} models the road by a meshgrid of surfels, which we adopt to generate pseudo-labels.

\noindent\textbf{Online mapping.}
Traditional HD maps constructed from survey vehicles \cite{ijgi13070232} are both expensive and prone to rapid obsolescence in dynamic environments. This has motivated the development of online mapping methods that generate maps in real time using vehicle-mounted sensors. Early approaches produce rasterized BEV segmentations \cite{zhou2022cross, li2022bevformer, philion2020lift} that lack the instance information required for tasks like motion planning \cite{hagedorn2024integration}, while others predict lane instances \cite{Pittner_2023_WACV,pittner2024lanecpp,luo2023latr} but do not include other map classes.

To overcome these limitations, vectorized map construction methods have emerged, starting with HDMapNet \cite{li2022hdmapnet}. VectorMapNet \cite{liu2023vectormapnet} reformulates online mapping as a detection task and thus adopts a one-to-one assignment between prediction and ground-truth elements as proposed by Detection Transformer (DETR) \cite{carion2020end}. MapTR \cite{liao2022maptr} further refined the assignment, and subsequent improvements were achieved through adaptations of the DETR-based queries, as seen in MapTRv2 \cite{liao2024maptrv2} and others \cite{choi2024mask2map,zhou2024himap}. Additional gains have been achieved by incorporating temporal context \cite{yuan2024streammapnet, chen2024maptracker}. MapVR \cite{zhang2024mapvr} introduces a differentiable rasterization loss as an auxiliary task, which we adapt for training with pseudo-labels.
While some work explores semi-supervised BEV segmentation \cite{lilja2024exploring,zhu2024semibevseg}, no semi- or unsupervised method has been proposed for vectorized online mapping. Our work aims to fill this gap.

\noindent\textbf{Offline mapping and annotation pipelines.}
Complementary to online mapping, offline mapping models learn to predict vectorized maps from temporal multi-view sensor data captured during single or multiple trips. After training the offline model with ground-truth maps, it can be deployed in data centers and automatically label large-scale crowdsourced data with further refinements by human annotators. The final maps serve as a more scalable alternative to traditional HD maps \cite{ xia2024dumapnet, xia2025ldmapnet}. MV-Map \cite{xie2023mv} learns offline mapping from single-trip data and ground-truth maps by incorporating a NeRF-based approach to enforce multi-view consistency. Building on the road reconstruction of RoMe \cite{mei2024rome}, CAMA \cite{zhang2024vision, chen2024camav2} generates novel BEV images that are processed by a BEV-compatible version of MapTR \cite{liao2022maptr} to predict a map. After refinements by human experts, this map is fed back into the model for additional training. A second branch of offline approaches stores historical maps onboard and uses them as priors during online mapping \cite{xiong2023neural, zhang2024enhancing, shi2024globalmapnet}.
Although we also propose an offline framework, \ours\ does not require ground-truth maps
and primarily trains a model to run online.

\section{Pseudo-Labels with Gaussian Splatting}
\label{sec:pseudo-labels}
The pseudo-labels are generated in four stages, which are shown in Fig.~\ref{fig:pipeline}. First, 2D semantic segmentation is performed.
Second, a 3D meshgrid of Gaussian surfels is initialized to model the color, semantics, and geometry of the road surface.
Third, an optimization procedure refines the surfel parameters by aligning the rendered outputs with both the raw camera images and the segmentation. Fourth, postprocessing techniques derive the vectorized map elements.

\subsection{Task Formulation}
Let the full set of images over a sequence of timestamps \( t_1, \dots, t_N \) be $\mathcal{I} = \{ I_c(t) \mid t \in \{t_n\}_{n=1}^N, c \in \mathcal{C} \},$ where \(\mathcal{C}\) denotes the set of available camera views. 
A pre-trained 2D semantic segmentation network \( f_\mathrm{seg} \) predicts a segmentation for each input image \( I \in \mathcal{I} \). 
Furthermore, the relative ego pose $e(t)$ and all sensor poses at timestamp $t$ are known.

Our goal is to generate a unified 3D road surface representation \(\mathcal{M}_\theta\) 
that explains the camera and segmentation observations across all views and timestamps. %
Once the optimized parameters \( \theta^* \) are obtained, a pseudo-label \( G(t) \) is generated by a bird's-eye view rendering $\mathrm{rend}_\mathrm{BEV}$ and a subsequent postprocessing $\mathrm{post}$:
\begin{equation}
    G(t) = \mathrm{post}\big(\mathrm{rend}_\mathrm{BEV}\big(e(t), \mathcal{M}_{\theta^\ast}\big)\big),
\end{equation}
where \( G(t) \) is the set of the final map elements for timestamp \( t \) represented as polyline or polygon vectors.

\subsection{Semantics}
Accurate semantic segmentation is crucial for generating high-quality pseudo-labels. To achieve this, we utilize a state-of-the-art segmentation model, Mask2Former \cite{cheng2022masked}, trained on the Mapillary Vistas V2 dataset \cite{neuhold2017mapillary}. We use its rich class taxonomy to remove segments such as arrows or text on the road surface.
Once trained, our segmentation network $f_\mathrm{seg}$ is deployed to infer semantic segmentations $\mathcal{I}_\mathrm{seg}=\{f_\mathrm{seg}(I)\}_{I\in\mathcal{I}}$. 
By shifting the labeling effort from costly HD map annotation to image segmentation, a well-established and scalable task, this approach significantly enhances the adaptability to diverse environments and is robust to different sensor arrangements that limit the generalizability of current online mapping models \cite{zhang2025mapgs}.

\subsection{Surface Model and Optimization}
In 3D Gaussian splatting \cite{kerbl20233d}, each Gaussian sphere can be modeled with a $3\times3$ rotation matrix $\mathbf{R}_\mathrm{GS}$ and a $3\times3$ scaling matrix $\mathbf{S}_\mathrm{GS}$. As recent work in surface reconstruction \cite{huang20242d} shows, reducing Gaussian spheres to flat surfels by restricting the scaling matrix to $\mathbf{S}_\mathrm{GS}=\mathrm{Diag}\left(\right[s_x, s_y, 0\left]\right)$ is more effective for modeling surfaces.

Therefore, we model the road as a meshgrid of Gaussian surfels $\mathcal{M}_\theta$, as illustrated in Fig.~\ref{fig:gaussian_surfels}. Following RoGs \cite{feng2024rogs}, we initialize the meshgrid along the vehicle poses $e(t)$ of the sequence with an offset $r$ in both the $x$ and $y$ direction. Each Gaussian surfel is parameterized by its 3D center coordinate, 3D orientation, 2D scale, opacity, color, and semantic class probability.

After initialization, the parameters $\theta$ of the Gaussian surfels are optimized by minimizing the discrepancy between rendered outputs, comprising both RGB and semantic channels, and the corresponding camera images $\mathcal{I}$ and PV segmentation labels $\mathcal{I}_\mathrm{seg}$. To ensure that only road-related information contributes to the optimization, we mask out pixels belonging to non-road classes, such as vehicles, buildings, and pedestrians, resulting in unobserved areas. Since road elements remain static over time, our approach does not require additional compensation for dynamic objects. Optionally, we can use LiDAR data to further improve the $z$-accuracy of each surfel.

\subsection{Postprocessing and Vectorization}

\begin{figure}
    \centering
    \includegraphics[clip, trim=6.5cm 2cm 7.5cm 1.5cm, width=0.75\linewidth]{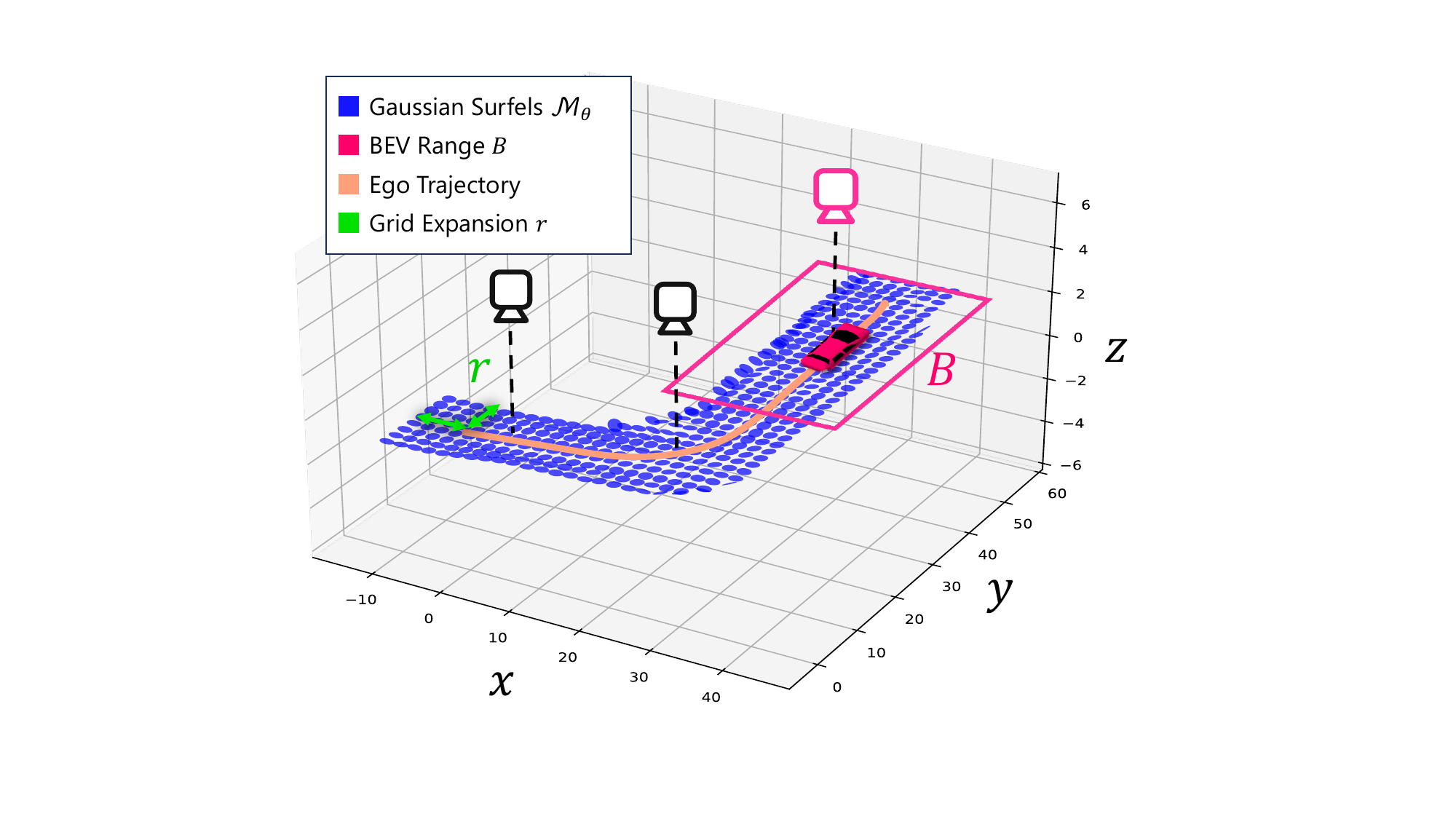}
    \caption{\textbf{Surface model and BEV rendering.} We initialize a 3D meshgrid $\mathcal{M}_\theta$ of flat Gaussian surfels along the vehicle's trajectory. After optimization, we render the orthographic BEV images for every vehicle pose $e(t)$.}
    \label{fig:gaussian_surfels}
\end{figure}

To ensure comparability with existing work in online mapping, we focus on the three most commonly used map classes: lane dividers, road boundaries, and pedestrian crossings, representing them as 2D polylines or polygons. Nonetheless, the approach is generalizable to any map class that can be reliably detected by a 2D segmentation network. Moreover, since our underlying road model is inherently 3D, our method could be extended to generate pseudo-labels for 3D online mapping.

\noindent\textbf{BEV rendering.} Once the surface optimization is finished, we render a semantic BEV segmentation map. To do this, for each timestamp \( t \), we place a virtual orthographic BEV camera at the \( xy \)-position of the vehicle pose \( e(t) \), oriented to match its heading, as illustrated in Fig.~\ref{fig:gaussian_surfels}. The camera is positioned to look directly downward along the negative \( z \)-axis, capturing the BEV range \( B \) on the \( xy \)-plane.

\noindent\textbf{Postprocessing.} Our postprocessing refines the initial BEV segmentation to produce vectorized map elements. Thereby, we remove small enclosed segments, followed by morphological filtering to remove spurious artifacts and to smooth segmentation boundaries. Fragmented lane markings are connected by dilation and then skeletonized. Road boundaries are extracted by the segment border between the road class and the adjacent outside classes, such as curbs, terrain, and driveways. All rasterized elements are then vectorized with an iterative procedure based on the Ramer-Douglas-Peucker algorithm \cite{ramer1972iterative, douglas1973algorithms}. We outline all postprocessing details and ablate the main parameters in Supp.~\ref*{sec:postprocessing}.

\subsection{Multi-trip Optimization}
As shown in Fig.~\ref{fig:pipeline}, the final pseudo map labels often exhibit large masked areas due to partial visits and occlusions. While our mask-aware assignment and loss function, introduced in Sec.~\ref{sec:training}, help to mitigate this issue, the potential of this method is limited as the assignment becomes more arbitrary with a higher mask ratio. Since our pseudo-label generation operates in an offline setting, we are not constrained to a single driving sequence. Instead, we can aggregate observations from multiple trips, potentially crowdsourced from fleet data. This approach not only enhances BEV coverage but also improves label consistency and quality, for instance, by supplementing nighttime sequences with data captured under daylight conditions.

Given multiple driving sequences with known relative poses in a common coordinate system, we initialize our meshgrid along the combined trajectories. The optimization of Gaussian surfels then proceeds in the same manner as for single trips but leverages a significantly larger set of observations, including more camera images, inferred PV segmentations, and LiDAR scans.

\section{Training with Pseudo-Labels}
\label{sec:training}

In vectorized online mapping, the objective is to predict a set of map elements \( Q \), represented as polygons or polylines, within a BEV range \( B \) using multi-view camera images \( \{I_c(t)\}_{c\in\mathcal{C}} \) at timestamp \( t \). In contrast, offline mapping uses the sensor measurements of an entire sequence.

For supervised approaches, the training loss is typically based on an optimal one-to-one assignment between a predicted and a ground-truth map element. The primary challenge in training an online mapping model with our pseudo-labels is the handling of incomplete and fragmented map elements since the generated labels suffer from occlusion by non-road objects and viewpoint limitations, as shown in Fig.~\ref{fig:o2m}. To address these challenges, we propose a mask-aware one-to-many assignment strategy and a corresponding loss function that accounts for the inherent uncertainty in the data and integrates it into the training process.

\subsection{Prediction Masking}
\label{sec:pred-masking}
\begin{figure}
    \centering
    \includegraphics[clip, trim=1.1cm 0.2cm 0.2cm 0.2cm, width=\linewidth]{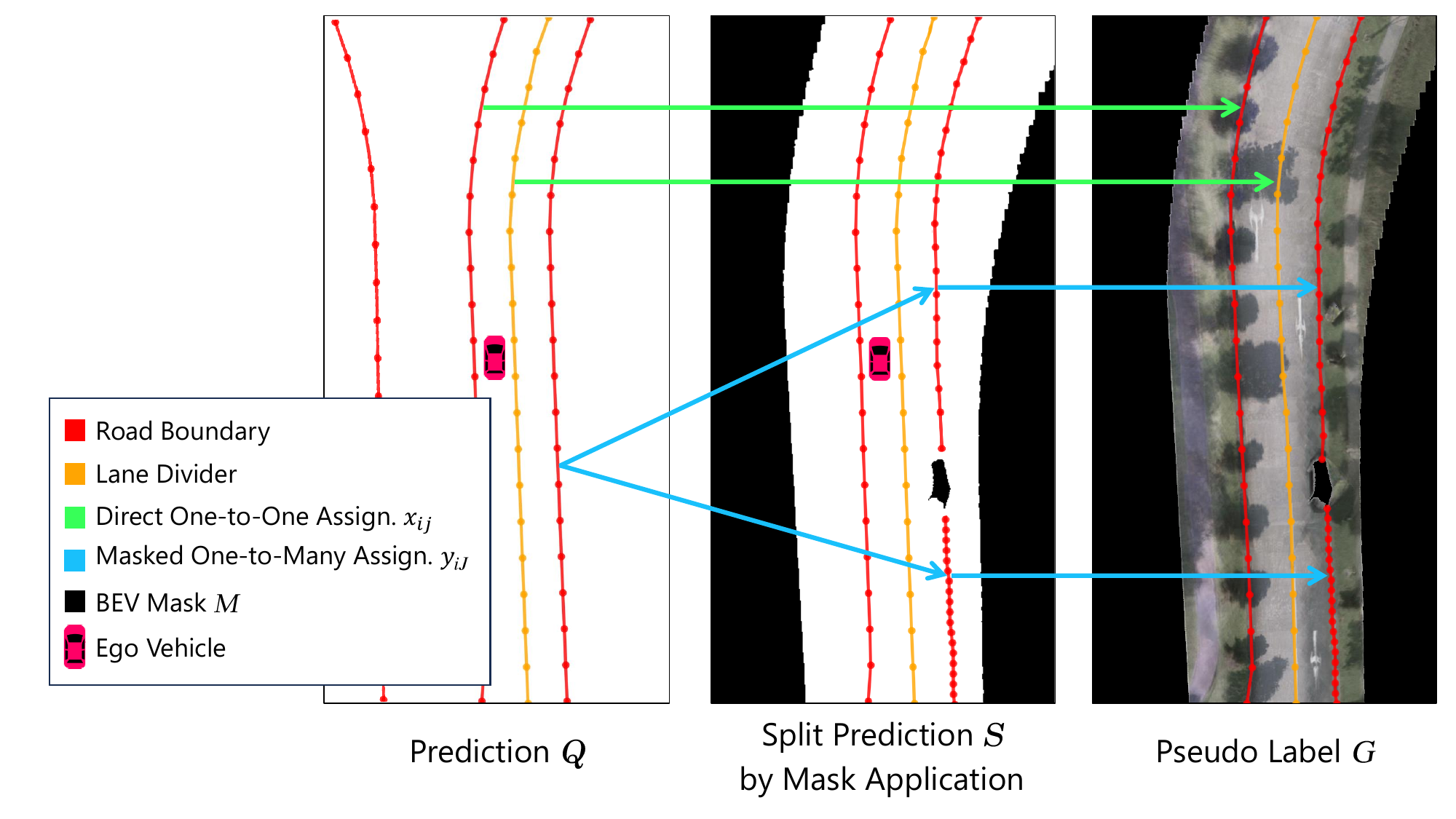}
    \caption{\textbf{One-to-many assignment.} In this example, one road boundary of the pseudo-label $G$ is partially interrupted by a parked car, which forms part of the BEV mask $M$. Thus, we propose a one-to-many assignment algorithm to match one predicted map element to many pseudo-label elements.}
    \label{fig:o2m}
\end{figure}

Let $Q=\{q_i\}_{i=1}^{|Q|}$ denotes the set of predictions of the online model and $G=\{g_j\}_{j=1}^{|G|}$ the set of pseudo-label elements represented as 2D vectors with $q_i,g_j\in\mathbb{R}^{L\times2}$.
For our mask-aware assignment and loss, we first need to apply the binary BEV mask $M$ to the prediction, as in Fig.~\ref{fig:o2m}. Thus, we $\mathrm{split}$ $q_i$ into a set of subsegments $S^i$ such that all points and their edges lie within the unmasked region:
\begin{equation}
    S^i = \left\{\mathrm{resample}(s, L) \mid s \in \mathrm{split}(q_i, M), |s| \geq L_\text{m} \right\}
\end{equation}
$\mathrm{resample}(s, L)$ standardizes a valid subsegment $s$ to a fixed length $L$ with subsegments shorter than $L_\text{m}$ are discarded. In practice, we choose $L_\text{m}=4$ since a lower value would lead to improper polygon resamples.

\subsection{Mask-aware Assignment}

To train an online model that predicts map elements across the full BEV range $B$, we introduce a hybrid assignment algorithm that combines two distinct strategies. 
First, we perform a standard one-to-one assignment, where each predicted element $q_i$ is matched to a single pseudo-label element $g_j$ based on the assignment cost $c_{\text{o2o}}(q_i, g_j)$. In this step, the BEV mask $M$ is not considered.
Second, to handle incomplete pseudo-labels, we introduce a mask-aware one-to-many assignment. 
In this case, a predicted element $q_i$, specifically its unmasked subsegments $S^i$, is matched to a subset of pseudo-label elements $J \subseteq G_\text{ind}$ using the cost $c_{\text{o2m}}(S^i, J)$. Here, $Q_\text{ind} = \{i\}_{i=1}^{|Q|}$ and $G_\text{ind} = \{j\}_{j=1}^{|G|}$ denote the sets of the corresponding indices.
The final assignment is obtained by solving a linear program.%

\noindent\textbf{Assignment costs.}
For the one-to-one assignment costs $c_{\text{o2o}}$, any common cost function can be chosen. We adopt the one by MapVR \cite{zhang2024mapvr} to be consistent with our loss selection as described in Sec.~\ref{sec:training-loss}. It adds a rendering cost to the class and position costs proposed by MapTR \cite{liao2022maptr}.

For the one-to-many assignment costs $c_{\text{o2m}}$, we collect the set $\mathcal{J}$ of all possible subsets of $G_\text{ind}$, where all corresponding elements of a subset $J\in\mathcal{J}$ belong to the same class. Now, the one-to-many cost is defined as:
\begin{equation} 
    c_{\text{o2m}}(S^i, J) = \begin{cases}
c_{\text{hungarian}}(S^i, J), & \text{ if } |S^i|=|J|\\
\infty, & \text{ otherwise}
\end{cases}
\end{equation}
with $ c_{\text{hungarian}}$ as the optimal cost based on the local matching $\pi\in\Pi_\mathrm{local}$ and the one-to-one cost function $c_{\text{o2o}}(q_i, g_j)$:
\begin{equation}
    c_{\text{hungarian}}(S^i, J) = \min_{\pi\in \Pi_\mathrm{local}}\sum_{j\in J} c_{\text{o2o}}(s^i_{\pi(j)},g_j)
\end{equation}
Here, the optimal local matching $\pi_\mathrm{local}^*$ between multiple pseudo-label elements and the prediction subsegments is found with the Hungarian algorithm \cite{kuhn1955hungarian}.

\noindent\textbf{Optimal assignment.} We solve the final global assignment with binary integer linear programming. It yields an optimal matching with $x^*_{ij},y^*_{iJ} \in \{0,1\}$ denoting the direct one-to-one assignment for $q_i$ and $g_j$ and the one-to-many assignment for $q_i$ and the corresponding elements in $J$, respectively. We summarize the assignment as
\begin{equation}
\pi_\mathrm{global}^*(i) =  \begin{cases}
\{j\}, & \text{ if } \exists j \text{ s.t. } x^*_{ij} =1\\
J, & \text{ if } \exists J \text{ s.t. } y^*_{iJ} =1\\
\emptyset, & \text{ otherwise}
\end{cases}.
\end{equation}
For the full problem formulation, we refer the reader to Supp. \ref*{sec:lp}.
In case that no prediction is split into more than one subsegment, i.e. $|S^i|\le1~\forall  q_i$, the problem can also be solved optimally with the Hungarian algorithm by padding $G$ with $\emptyset$ elements such that $|G|=|Q|$. This makes the assignment faster during training.

Note that our one-to-many assignment differs from the one used in MapTRv2~\cite{liao2024maptrv2}, where a single ground-truth element is assigned to multiple predicted auxiliary elements. In contrast, we assign one predicted element to zero, one, or several pseudo-ground-truth elements. Both approaches are complementary, but we exclude the one from MapTRv2 as it dramatically increases the combinatorial complexity of the assignment. While this can be handled by the Hungarian algorithm in MapTRv2, it would increase training times by an order of magnitude when used with our linear program solver, making it impractical.

\subsection{Mask-aware Loss}
\label{sec:training-loss}

We build upon the losses proposed by MapTRv2~\cite{liao2024maptrv2} and MapVR~\cite{zhang2024mapvr}, extending them to handle partially masked predictions. Predicted elements that are completely masked and also not matched one-to-one are excluded from the loss. The remaining elements form the subset $Q^\prime_\mathrm{ind}\subseteq Q_\mathrm{ind}$ used for calculating the final loss.

\noindent\textbf{Classification loss.} 
Given the optimal assignment $\pi_\mathrm{global}^*(i)$, the classification loss is defined using the Focal loss~\cite{lin2017focal} with the predicted class probability $\hat{p}_i$ and the class label $\mathrm{cls}$ of the assigned pseudo-label element:
\begin{equation}
\mathcal{L}_{\mathrm{cls}}=\sum_{i\in Q^\prime_\text{ind}} \mathcal{L}_{\text {Focal }}\left(\hat{p}_{i}, \mathrm{cls}(\pi_\mathrm{global}^*(i))\right)
\end{equation}

\noindent\textbf{Point-wise loss.}  
For a one-to-many matching, a point-wise L1 loss as in MapTR \cite{liao2022maptr} is not straightforward since a single predicted map element may correspond to multiple pseudo-label elements. Thus, we compute the loss exclusively for direct one-to-one assignments where \( x_{ij}^* = 1 \):
\begin{equation}
    \mathcal{L}_{\mathrm{pt}} = \sum_{i\in Q^\prime_\text{ind}} \sum_{j\in G_\text{ind}} x_{ij}^* \sum_{l=1}^{L} \|q_{i,l} - g_{j,\gamma_j(l)}\|_1.
\end{equation}
\( \gamma_j(l) \) denotes the optimal point-wise assignment for each predicted point to its corresponding pseudo-label point.

\noindent\textbf{Rendering loss.}
MapVR introduces a differentiable rendering loss, where each map element is first rasterized, and then the Dice loss \cite{milletari2016v} is computed between prediction and ground-truth rasterizations.

We find this loss particularly well suited for adaptation to our one-to-many assignment as we can render all pseudo-label elements $\{g_j\}_{j\in\pi_\mathrm{global}^*(i)}$ assigned to a single prediction $q_i$ into a unified raster. The Dice loss is then computed between this aggregated rasterization and the rasterized prediction of $q_i$. Additionally, we apply the BEV mask $M$ to exclude unobserved regions, ensuring that the loss is computed only over unmasked grid cells. This rendering loss, $\mathcal{L}_\mathrm{rend}$, serves as an effective alternative to the point-wise loss for one-to-many assignments.

\noindent\textbf{Direction loss.} We adopt the self-supervised direction loss $\mathcal{L}_\mathrm{dir}$ from MapVR, which regularizes the model and prevents overfitting to imperfect pseudo-labels.

\begin{figure}
    \centering
    \begin{subfigure}[b]{0.49\linewidth}
        \centering
        \includegraphics[width=\linewidth]{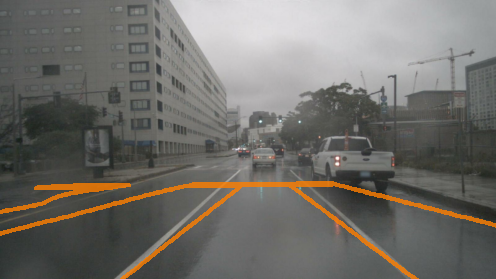}
        \caption{Projected GT in MapTRv2 \cite{liao2024maptrv2}}
        \label{fig:pv_seg_maptr}
    \end{subfigure}
    \hfill
    \begin{subfigure}[b]{0.49\linewidth}
        \centering
        \includegraphics[width=\linewidth]{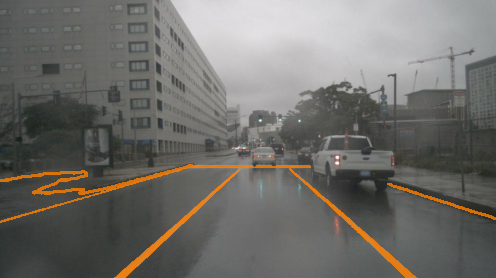}
        \caption{Our PV segmentation}
        \label{fig:pv_seg_ours}
    \end{subfigure}
    \caption{\textbf{PV segmentation.} The projected ground-truth (GT) map in MapTRv2 suffers from misalignment, while our PV segmentation aligns well as it is derived directly from the image. This provides more precise supervision for the PV features.}
    \label{fig:pv_seg}
\end{figure}

\noindent\textbf{Segmentation loss.}
Following MapTRv2, we adopt the binary segmentation loss in BEV and perspective view (PV) for auxiliary supervision of the BEV and PV features. Analogous to the rendering loss, we mitigate the impact of unobserved regions by applying the mask $M$ to the BEV predictions before calculating the final loss $\mathcal{L}_\mathrm{BEV}$.

For the PV segmentation loss $\mathcal{L}_\mathrm{PV}$, MapTRv2 projects and rasterizes the ground-truth map to supervise the PV features. Since some datasets like nuScenes \cite{caesar2020nuscenes} contain only 2D maps without elevation, the projected map will be misaligned with the actual image, as demonstrated in Fig.~\ref{fig:pv_seg_maptr}.

Given that we have direct access to high-quality PV image segmentation produced by our pre-trained segmentation network $f_\mathrm{seg}$, we can leverage this data to produce more accurate labels. In comparison to our map pseudo-labels, the PV segmentation has not undergone the aforementioned postprocessing steps, where some of the information naturally gets lost. The PV segmentation contains particularly valuable information and is also unrestrictedly available compared to the masked BEV. Thus, we extract the map segments of our original segmentation images $\mathcal{I}_\mathrm{seg}$ and downsample them to the dimension of the PV feature map using max-pooling. We notice that our segmentations are more aligned with the actual image than the projected 2D ground-truth map utilized in MapTRv2, as shown in Fig.~\ref{fig:pv_seg}.

\noindent\textbf{Depth and final loss.} 
We adopt the depth loss $\mathcal{L}_\mathrm{depth}$ from MapTRv2~\cite{liao2024maptrv2}, leveraging LiDAR data for additional depth supervision during training. However, the model itself remains camera-only.
The final loss $\mathcal{L}$ is a weighted sum of the losses $\mathcal{L}_{\mathrm{cls}}$, $\mathcal{L}_{\mathrm{pt}}$, $\mathcal{L}_\mathrm{rend}$, $\mathcal{L}_{\mathrm{dir}}$, $\mathcal{L}_{\mathrm{BEV}}$, $\mathcal{L}_{\mathrm{PV}}$, and $\mathcal{L}_\mathrm{depth}$.

\section{Experiments}
\label{sec:exp}
\subsection{Experimental Setup}

\noindent \textbf{Dataset.} We conduct our experiments on nuScenes~\cite{caesar2020nuscenes}, a large-scale dataset that provides multi-view images, LiDAR, and HD map annotations. As shown by multiple studies \cite{yuan2024streammapnet, lilja2024localization}, the original training and validation set contain highly overlapping locations, such that results reported on this split demonstrate memorization rather than generalization capabilities. Thus, we train and evaluate all models on the geographically disjoint data split proposed by Lilja et al. \cite{lilja2024localization}.
Furthermore, we select the three main map classes, such as lane dividers, road boundaries, and pedestrian crossings, for evaluation. The pseudo-labels are generated from single and multiple trips, in both cases using LiDAR data to supervise the elevation of the road surface.

\noindent \textbf{Metrics.} 
We adopt the average precision (AP) based on the Chamfer distance as introduced by HDMapNet \cite{li2022hdmapnet} with common thresholds of \{\SI{0.5}{m}, \SI{1.0}{m}, \SI{1.5}{m}\} and report the average for all map classes. In addition, we report the inference speed in frames per second (FPS).

\noindent \textbf{Implementation details.}
All evaluated online models utilize a camera-only sensor setup with a ResNet-50 \cite{he2016deep} image backbone. All details regarding the pseudo-label generation can be found in Supp.~\ref*{sec:appendix-exp-details}.

 \begin{figure}
    \centering
    \includegraphics[width=\linewidth]{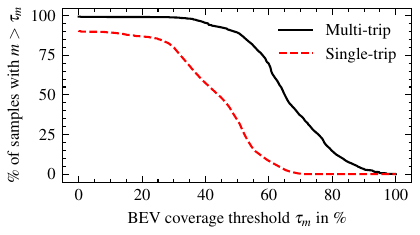}
    \caption{\textbf{BEV coverage evaluation.} Comparison of the BEV coverage $\covratio$ between pseudo-labels generated from a single trip and multiple trips on the training set.}
    \label{fig:bev-coverage}
\end{figure}

\subsection{How accurate are the pseudo-labels?}
\label{sec:pseudo-eval}
We evaluate the quality of our pseudo-labels by their BEV coverage and accuracy compared to ground-truth HD maps.

\noindent\textbf{BEV coverage.}
Since the pseudo-labels do not cover the entire BEV range $B$, we evaluate the coverage ratio, denoted as $\covratio$, which measures the proportion of the unmasked BEV range. We plot the percentage of training samples that exceed a given threshold $\tau_{\covratio}$ in Fig.~\ref{fig:bev-coverage}. For single-trip data, we achieve an average coverage ratio of \SI{40.0}{\%}. Extending to multi-trip data increases this ratio to \SI{65.5}{\%}, with almost all samples exceeding \SI{30}{\%} coverage. These results highlight the potential of aggregating crowdsourced data, where multiple vehicles contribute partial observations to construct a more complete map.

\let\hline\midrule %

\begin{table}[t]
\caption{\textbf{Offline performance.} Evaluation of the pseudo-labels on the validation set based on the complete BEV range or on the observed area only.}
\centering
\resizebox{\linewidth}{!}{
\begin{tabular}{c|l|S[table-format=2.1]S[table-format=2.1]S[table-format=2.1]>{\columncolor{gray!20}}S[table-format=2.1]}
\toprule
Pseudo-label & \multicolumn{1}{c|}{\multirow{2}{*}{Evaluation area}} & \multicolumn{4}{c}{AP} \\
 collection        & \multicolumn{1}{c|}{}                                    & \multicolumn{1}{l}{ped.} & \multicolumn{1}{c}{div.} & \multicolumn{1}{c}{bdry.} & \multicolumn{1}{>{\columncolor{gray!20}}c}{\textit{mean}}\\ 
 \midrule

\multirow{2}{*}{\begin{tabular}[c]{@{}c@{}}Single-trip\end{tabular}} & Full BEV range & 9.5 & 1.9  & 3.2  & 4.9  \\
 & Observed area only   & 23.6 & 6.0  & 26.8 & 18.8 \\
 \midrule

\multirow{2}{*}{\begin{tabular}[c]{@{}c@{}}Multi-trip\end{tabular}} & Full BEV range  & 18.3 & 1.9  & 10.7 & 10.3 \\
  & Observed area only   & 25.8 & 2.3  & 26.2 & 18.1 \\
\bottomrule
\end{tabular}
}
\label{tab:pseudo-eval}
\end{table}

\begin{figure*}[t]
    \centering
    \begin{subfigure}[b]{0.103\linewidth}
        \centering
        \includegraphics[width=\linewidth]{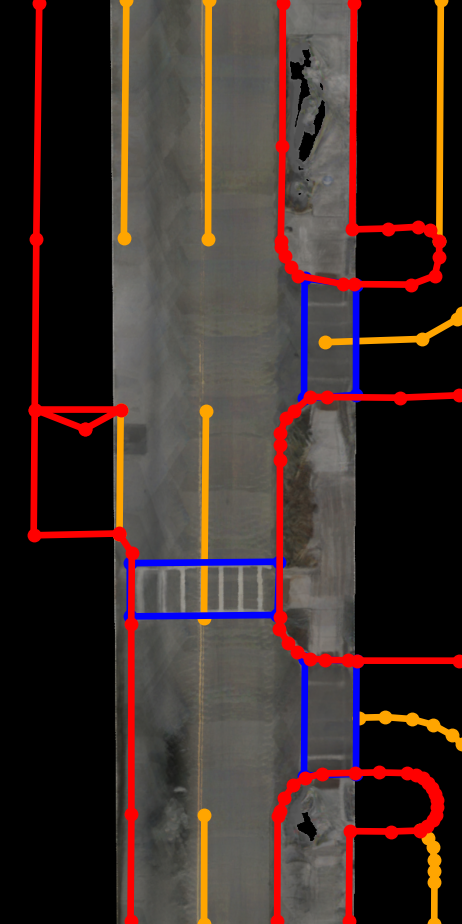}
        \caption{GT}
        \label{fig:1-gt}
    \end{subfigure}
    \begin{subfigure}[b]{0.103\linewidth}
        \centering
        \includegraphics[width=\linewidth]{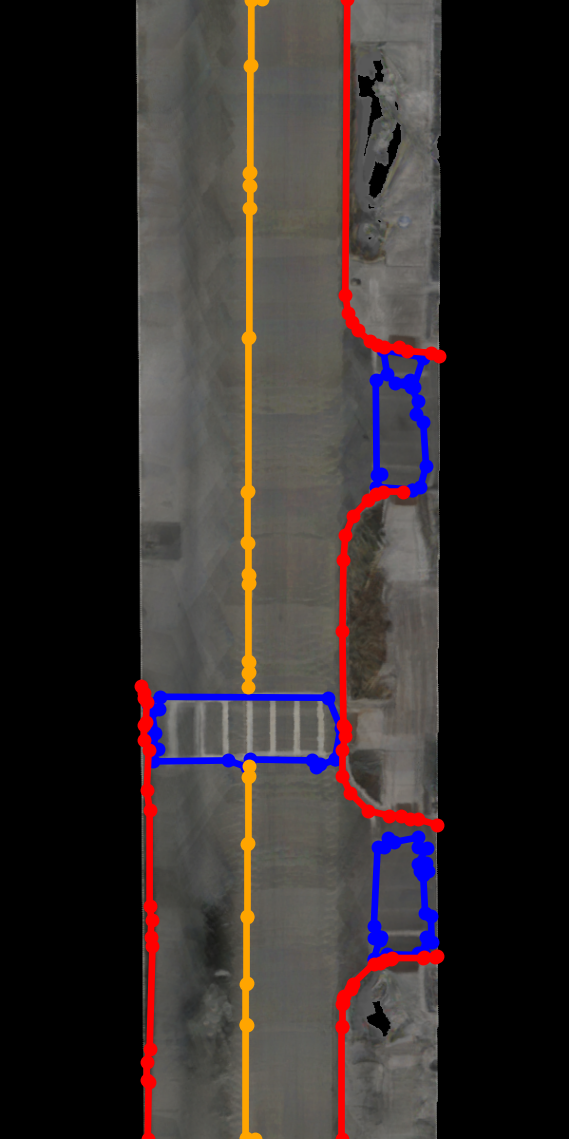}
        \caption{Single-trip}
        \label{fig:1-single}
    \end{subfigure}
    \begin{subfigure}[b]{0.103\linewidth}
        \centering
        \includegraphics[width=\linewidth]{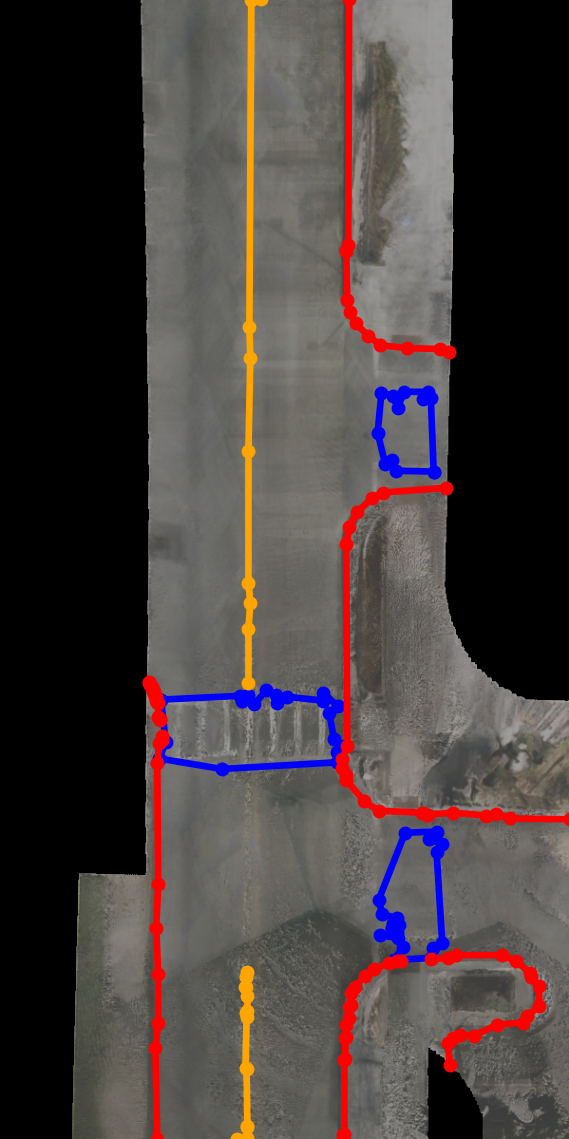}
        \caption{Multi-trip}
        \label{fig:1-multi}
    \end{subfigure}
    \hfill
    \begin{subfigure}[b]{0.103\linewidth}
        \centering
        \includegraphics[width=\linewidth]{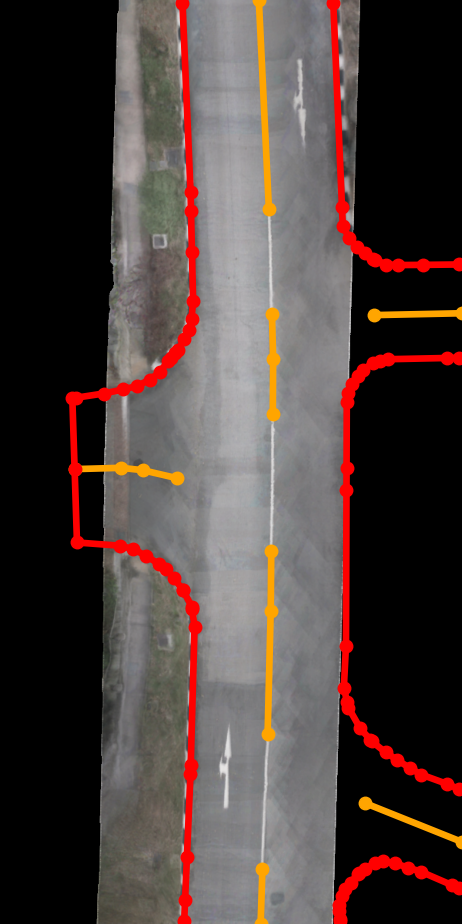}
        \caption{GT}
        \label{fig:2-gt}
    \end{subfigure}
    \begin{subfigure}[b]{0.103\linewidth}
        \centering
        \includegraphics[width=\linewidth]{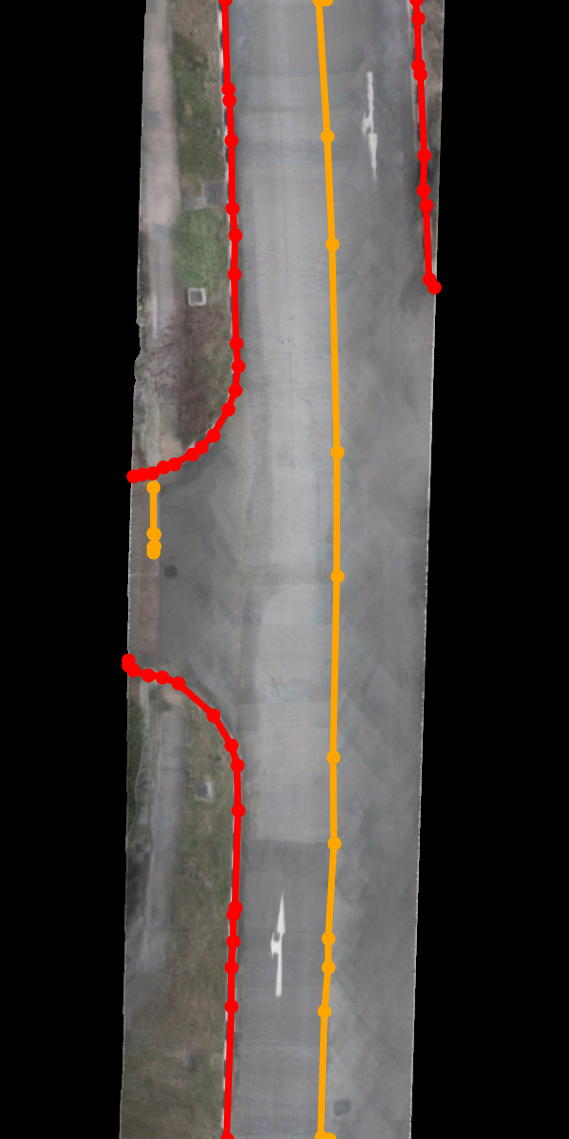}
        \caption{Single-trip}
        \label{fig:2-single}
    \end{subfigure}
    \begin{subfigure}[b]{0.103\linewidth}
        \centering
        \includegraphics[width=\linewidth]{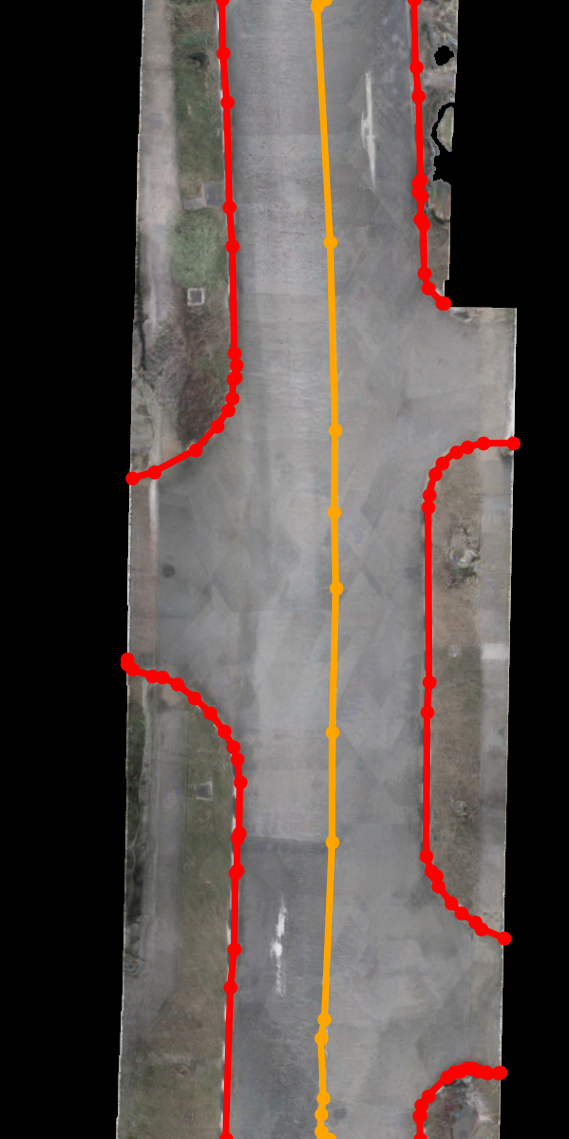}
        \caption{Multi-trip}
        \label{fig:2-multi}
    \end{subfigure}
    \hfill
    \begin{subfigure}[b]{0.103\linewidth}
        \centering
        \includegraphics[width=\linewidth]{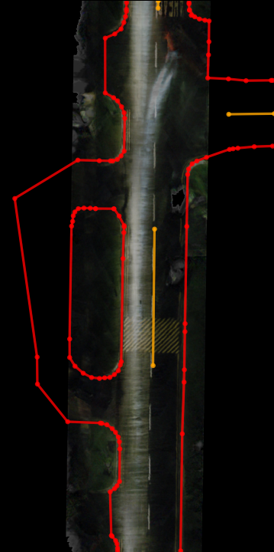}
        \caption{GT}
        \label{fig:3-gt}
    \end{subfigure}
    \begin{subfigure}[b]{0.102897\linewidth}
        \centering
        \includegraphics[width=\linewidth]{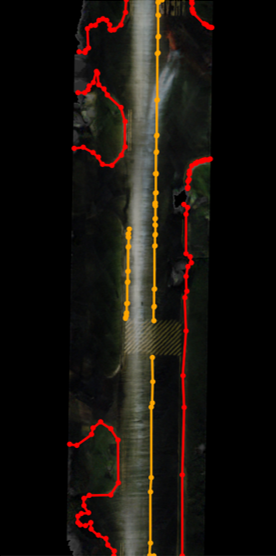}
        \caption{Single-trip}
        \label{fig:3-single}
    \end{subfigure}
    \begin{subfigure}[b]{0.103206\linewidth}
        \centering
        \includegraphics[width=\linewidth]{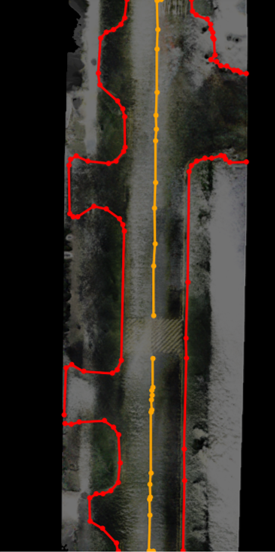}
        \caption{Multi-trip}
        \label{fig:3-multi}
    \end{subfigure}
    \caption{\textbf{Qualitative pseudo-label evaluation.} Comparison of the generated pseudo-labels from a single trip and multiple trips with the ground truth (GT) for three diverse scenes with (g)-(i) showing a low-light scenario. We plot the lane dividers (orange), road boundaries (red), and pedestrian crossings (blue) and use the colored BEV rendering as background for a visual evaluation. In all three cases, we identify inconsistencies for the ground-truth lane dividers, where the pseudo-labels sometimes provide more plausible results. Additional qualitative results are provided in Supp.~\ref*{sec:appendix-qualitative}.}
    \label{fig:qualitative}
\end{figure*}

\noindent\textbf{Comparison with ground truth.}  
To assess the quality of our pseudo-labels, we compare them against the ground-truth map of the validation set in Tab.~\ref{tab:pseudo-eval}. We conduct evaluations under two conditions: (1) comparing pseudo-labels against all ground-truth elements, including those in unobserved regions, and (2) restricting the evaluation to only the observed regions by applying the BEV mask $M$ to the ground truth, similar to the prediction masking in Sec.~\ref{sec:pred-masking}. The latter provides a more precise assessment of the accuracy in the areas covered by the pseudo-labels. In addition, we provide qualitative results in Fig.~\ref{fig:qualitative} as well as in Supp.~\ref*{sec:appendix-qualitative}.

The results in Tab.~\ref{tab:pseudo-eval} highlight substantial differences across map classes. Road boundaries, typically located farther from the vehicle's trajectory, are often underrepresented in the pseudo-labels, resulting in lower performance when evaluated across the full BEV range. The lane divider class exhibits particularly poor performance, which can partially be attributed to inconsistencies in the ground-truth annotations, as shown in Fig.~\labelcref{fig:1-gt,fig:2-gt,fig:3-gt} and noted in previous work \cite{chen2024camav2,zhang2024vision}. Also, lane markings, being narrow structures, are more prone to being overridden by adjacent road class segments during optimization, causing them to disappear in the final BEV segmentation. This issue arises when camera poses are suboptimal, which is especially common in multi-trip data, see Fig.~\ref{fig:1-multi}. This also explains the lower performance for lane dividers coming from multiple trips.
Despite these limitations, the pseudo-labels for pedestrian crossings and road boundaries within observed areas achieve comparable performance to the best-performing online models in Tab.~\ref{tab:main} trained on ground truth.

\begin{table*}[t]
    \caption{\textbf{Online mapping performance.} Comparison of our method and baselines on the validation set trained on ground truth or pseudo-labels. We highlight the best results per type of training label. $\dagger$ means the results reported by \cite{lilja2024localization}. The FPS results are taken from MapTRv2~\cite{liao2024maptrv2}. $\ast$ indicates methods that have access to previous frames for prediction.}
    \centering
    \begin{tabular}{cl|l|c|S[table-format=2.1]|S[table-format=2.1]S[table-format=2.1]S[table-format=2.1]>{\columncolor{gray!20}}S[table-format=2.1]}
    \toprule
    \multicolumn{2}{c|}{\multirow{2}{*}{Training Labels}}  & \multicolumn{1}{c|}{\multirow{2}{*}{Method}}  & \multirow{2}{*}{Epochs} & \multicolumn{1}{c|}{\multirow{2}{*}{FPS}} & \multicolumn{4}{c}{AP} \\
     & & \multicolumn{1}{c|}{} &  & & \multicolumn{1}{l}{ped.} & \multicolumn{1}{c}{div.} & \multicolumn{1}{c}{bdry.} & \multicolumn{1}{>{\columncolor{gray!20}}c}{\textit{mean}} \\ 
    \midrule
    \multicolumn{2}{l|}{\multirow{5}{*}{\textbf{Ground Truth}}}
       & VectorMapNet \cite{liu2023vectormapnet}$^{\dagger}$ & 110 & 2.2  & 13.7 & 13.5 & 14.9 & 14.0 \\
      & & MapTR \cite{liao2022maptr}$^{\dagger}$       & 24  & 15.1  & 14.4 & 16.0 & 26.7 & 19.0  \\
      & & MapVR \cite{zhang2024mapvr}       & 24  & 15.1  & 17.0 & 16.3 & 27.6 & 20.3  \\
      & & StreamMapNet \cite{yuan2024streammapnet}$^{\dagger}$$^{\ast}$ & 24  & \multicolumn{1}{c|}{--}  & \textbf{25.8} & \textbf{23.0} & 29.5 & \textbf{26.1}  \\
      & & MapTRv2 \cite{liao2024maptrv2}               & 24  & 14.1  & 23.8 & 19.5 & \textbf{32.7} & 25.4  \\
    \midrule
    \multirow{4}{*}{\begin{tabular}[l]{@{}l@{}}\textbf{Pseudo-}\\ \textbf{Labels}\end{tabular}} 
    
    & \multirow{2}{*}{Single-trip}
       & MapTRv2 \cite{liao2024maptrv2} & 24  & 14.1  & 9.9  & 3.2  & 7.0  & 6.7  \\
      & & MapTRv2 \cite{liao2024maptrv2} + \ours\  & 24  & 14.1  &  \textbf{12.3} & \textbf{~~3.8} & \textbf{~~8.3}  & \textbf{~~8.2}  \\
    \cmidrule(lr){2-9}
    
      & \multirow{2}{*}{Multi-trip} & MapTRv2 \cite{liao2024maptrv2} & 24  & 14.1  & 14.4 & 2.6  & 14.5 & 10.5 \\
      &  & MapTRv2 \cite{liao2024maptrv2} + \ours\ & 24  & 14.1  & \textbf{18.1} & \textbf{~~4.1} & \textbf{17.4} & \textbf{13.2} \\
    \bottomrule
    \end{tabular}
    \label{tab:main}
\end{table*}

\subsection{Can online models train on pseudo-labels?}
\noindent\textbf{Main results.}  
We compare our approach against common supervised baseline methods in Tab.~\ref{tab:main}. For MapTRv2 \cite{liao2024maptrv2}, we conduct additional experiments by training the model naively on pseudo-labels from single and multiple trips. To ensure meaningful training, we filter out samples with a BEV coverage below $\tau_{\covratio}=\SI{50}{\%}$. When training MapTRv2 with \ours, we lower the coverage threshold to $\tau_{\covratio}=\SI{30}{\%}$ for single-trip training as it can handle unobserved areas due to its mask-aware approach.

As expected, MapTRv2 trained on pseudo-labels exhibits a significant performance drop, particularly when using single-trip data. However, incorporating multiple trips improves performance, benefiting from more consistent pseudo-labels and increased BEV coverage. Further gains are achieved when training MapTRv2 with \ours, which outperforms VectorMapNet \cite{liu2023vectormapnet} on pedestrian crossings and boundary elements - despite never being trained on ground truth. Nonetheless, compared to the best-performing supervised methods, our approach still has a notable performance gap, especially for the lane divider class. This shortfall is attributed to the differences between ground truth and pseudo-labels, as discussed in Sec.~\ref{sec:pseudo-eval}.

\begin{table}[t]
    \caption{\textbf{Ablation study.} Performance comparison of key components of \ours, trained on single-trip pseudo-labels.}
    \centering
    \resizebox{\linewidth}{!}{
    \begin{tabular}{l|S[table-format=2.1]S[table-format=1.1]S[table-format=1.1]>{\columncolor{gray!20}}S[table-format=1.1]}
    \toprule
    \multicolumn{1}{c|}{\multirow{2}{*}{Training Configuration}} & \multicolumn{4}{c}{AP} \\
    \multicolumn{1}{c|}{} & \multicolumn{1}{l}{ped.} & \multicolumn{1}{c}{div.} & \multicolumn{1}{c}{bdry.} & \multicolumn{1}{>{\columncolor{gray!20}}c}{\textit{mean}} \\ 
    \midrule
    Baseline (MapTRv2, $\tau_{\covratio}=\SI{50}{\%}$) & 9.9  & 3.2  & 7.0  & 6.7 \\
    + lower $\tau_{\covratio}$ to $\SI{30}{\%}$ & 8.4 & 2.6 & 5.0 & 5.3 \\
    + rendering \& direction losses & 10.9 & 3.5 & 4.0 & 6.1  \\
    + PV segmentation loss w/o projection & 10.9 & 3.6 & 4.7 & 6.4 \\
    + mask-aware assignment \& loss & \textbf{12.3} & \textbf{3.8} & \textbf{8.3} & \textbf{8.2} \\
    \bottomrule
    \end{tabular}
    }
    \label{tab:ablations}
\end{table}

\noindent\textbf{Ablation study.}  
We conduct an ablation study on the key components of \ours, training on pseudo-labels from single trips. The results are summarized in Tab.~\ref{tab:ablations}. A naive training of MapTRv2 struggles with pseudo-labels when the BEV coverage falls below $\SI{50}{\%}$. Introducing the rendering and directional losses proposed by MapVR \cite{zhang2024mapvr} along with the PV segmentation labels derived directly from our segmentation network, leads to slight performance improvements. A significant gain is observed when incorporating the mask-aware assignment and loss, which enables the model to effectively leverage low-coverage samples. This is particularly evident for boundary elements, which are mostly located in unobserved regions.%

\subsection{How does it benefit semi-supervised learning?}
\begin{figure}[t]
    \centering
    \includegraphics[width=0.9\linewidth]{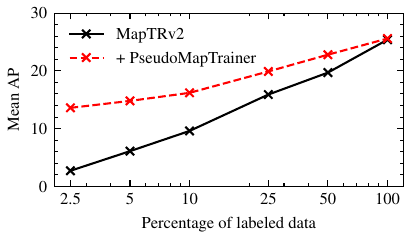}
    \caption{\textbf{Semi-supervised training.} 
    Performance comparison between supervised MapTRv2 training and the same model pre-trained on pseudo-labels.}
    \label{fig:semi-super}
\end{figure}

\ours\ can be used to pre-train a model that is later fine-tuned with ground-truth maps in a semi-supervised manner. To evaluate its effectiveness, we pre-train MapTRv2 with \ours\ using multi-trip pseudo-labels from the full training set and then fine-tune it on a fraction of the available ground-truth labels. The performance is compared to a purely supervised MapTRv2 baseline in Fig.~\ref{fig:semi-super}. Our results show that \ours\ significantly improves performance, particularly in low-label regimes. This highlights its potential for enhancing online mapping in large-scale scenarios where abundant unlabeled data, such as crowdsourced data, is available, but ground-truth annotations are limited.

\subsection{Limitations}
\label{sec:limitations}
Like most camera-based methods, our offline mapping approach is sensitive to challenging lighting conditions, such as nighttime (see Fig.~\ref{fig:3-single}). However, incorporating data from multiple trips under different conditions helps mitigate these limitations, as demonstrated in Fig.~\ref{fig:3-multi}.

Merging the data from multiple trips requires highly precise relative positioning between sequences, which we presuppose in this study. In practice, achieving such precision can be challenging, particularly for vehicles relying solely on cameras and consumer-grade positioning systems. 
However, using additional LiDAR or radar sensors, previous work \cite{Löwens_2024_BMVC, liu2024extend} showed that the relative vehicle poses can be accurately recovered based on unsupervised learned registration methods. 
Additionally, care must be taken to ensure that merged sequences correspond to timestamps without significant road changes, such as construction. Thus, we propose both a single-trip and a multi-trip approach for pseudo-label generation, providing flexibility depending on sensor availability and environmental stability.

\section{Conclusion}
We demonstrate the effectiveness of training online mapping models without relying on ground-truth HD maps. Our pseudo-labels also enable efficient pre-training in semi-supervised scenarios with significant performance improvements. This highlights the value of leveraging large-scale crowdsourced data for scalable online mapping.

However, we still see potential for future work. In particular, the lane dividers need to be better preserved through targeted adaptations of the Gaussian optimization process. In addition, incorporating inexpensive SD maps and satellite images could further improve the pseudo-label quality. Another promising direction are pseudo-labels for centerlines, as discussed in Supp.~\ref*{sec:appendix-centerlines}. For the online model training, self-supervised pre-training presents an opportunity to improve robustness against noisy pseudo-labels.

{
    \small
    \bibliographystyle{ieeenat_fullname}
    \bibliography{main}
}
\clearpage
\setcounter{page}{1}
\appendix
\renewcommand{\thefigure}{S.\arabic{figure}}
\renewcommand{\thetable}{S.\arabic{table}}
\setcounter{figure}{0}
\setcounter{table}{0}
\maketitlesupplementary

\begin{figure}
    \centering
    \begin{subfigure}[b]{0.3\linewidth}
        \centering
        \includegraphics[width=\linewidth]{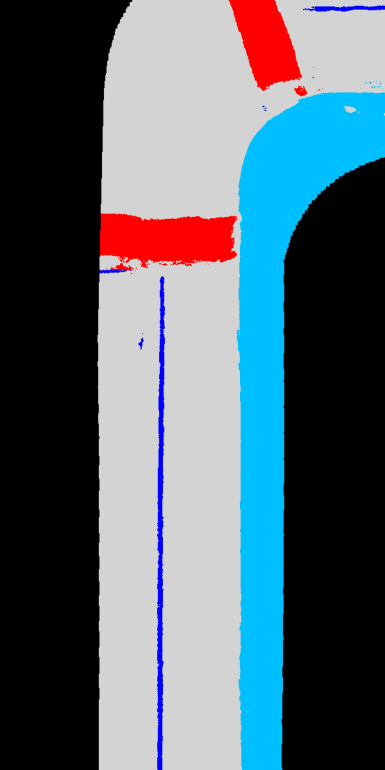}
        \caption{BEV rendering}
        \label{fig:postraw}
    \end{subfigure}
    \hfill
    \begin{subfigure}[b]{0.3\linewidth}
        \centering
        \includegraphics[width=\linewidth]{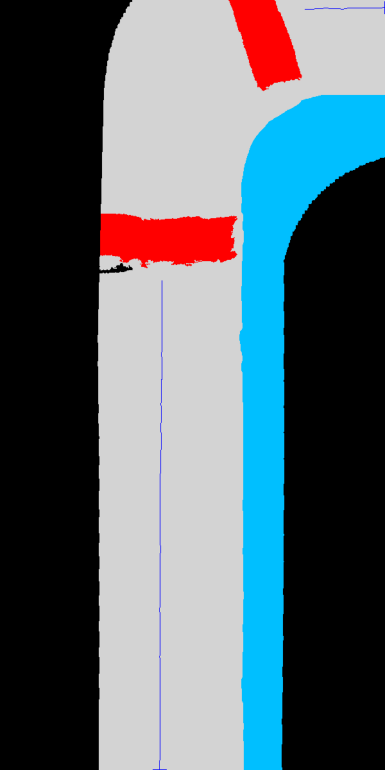}
        \caption{Smoothed}
        \label{fig:postpost}
    \end{subfigure}
    \hfill
    \begin{subfigure}[b]{0.3\linewidth}
        \centering
        \includegraphics[width=\linewidth]{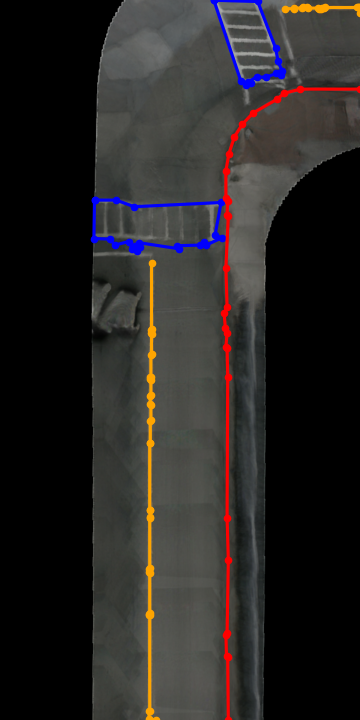}
        \caption{Vectorization}
        \label{fig:postvect}
    \end{subfigure}
    \caption{\textbf{Postprocessing and vectorization.} We remove artifacts in the raw semantic BEV rendering and further smooth the road boundary for accurate polyline and polygon extraction.}
    \label{fig:postprocessing}
\end{figure}

\section{Postprocessing}
\label{sec:postprocessing}
Our postprocessing pipeline refines the initial BEV segmentation to address common artifacts introduced by the surface reconstruction and generates vectorized map elements suitable for training, as shown in Fig.~\ref{fig:postprocessing}. In practice, we extend the BEV renderings by a small margin before postprocessing it to avoid boundary effects. After we obtain the vectorized elements, we crop them to the desired range.

\subsection{Overall pipeline}
\label{sec:postprocessing-pipe}
The postprocessing pipeline consists of the following steps:

\noindent\textbf{Removing artifacts.} 
Due to inaccuracies in the surfel optimization, small segments can be misclassified. To reduce these artifacts, we employ a class-based connected-component labeling \cite{rosenfeld1966sequential} to identify small segments enclosed by other segments. These are then reassigned to the enclosing classes to ensure semantic consistency. Small segments that are adjacent to more than one class are removed by assigning them either to the BEV mask or to one of the adjacent classes. We also remove lane-marking segments (dark blue in Fig.~\ref{fig:postraw}) that are unreasonably thick.

\noindent\textbf{Extracting the road boundary.}
We apply morphological filtering to the outside class (light blue in Fig.~\ref{fig:postprocessing}), resulting in a smoother road boundary, which is extracted by the border between segments of the road class and segments of the outside class.

\noindent\textbf{Vectorization of line-shaped elements.} 
We connect spatially close lane-marking fragments through dilation to become the lane dividers. We skeletonize the lane divider and road boundary segments using the Zhang-Suen algorithm \cite{zhang1984fast} into line components. For Y-shaped lines, the longest path is preserved, while the other branches become new components. The lines are subsequently converted into polylines and simplified through iterative polygonal approximation based on the Ramer-Douglas-Peucker (RDP) algorithm \cite{douglas1973algorithms}. We initialize the maximum distance threshold with $\epsilon^{(1)}$ and iteratively increase it as $\epsilon^{(t)} = \epsilon^{(1)}t$ until the simplified polyline contains no more than $L$ points.
Finally, lane dividers that are overly close and parallel to boundaries or pedestrian crossings are removed.

\noindent\textbf{Vectorization of polygon-shaped elements.} To extract the borders of pedestrian crossings, we first employ the Suzuki-Abe border-following algorithm \cite{suzuki1985topological}. Similar to line-shaped elements, we then apply the RDP algorithm to obtain a simplified yet accurate polygonal representation.

\begin{table}[t]
    \caption{Pipeline ablation for the observed region in single trips. The default parameters are: \SI{20}{pixel/m}, 15, and \SI{5}{cm}.}
    \scriptsize
    \centering
    
    \setlength{\tabcolsep}{4pt}
    \begin{tabular}{
        l
        |S[table-format=2.1]
        |S[table-format=2.1]S[table-format=2.1]S[table-format=2.1]
        |S[table-format=2.1]S[table-format=2.1]S[table-format=2.1]
        |S[table-format=2.1]S[table-format=2.1]S[table-format=2.1]
    }
    \toprule
        \multicolumn{1}{c|}{\multirow{2}{*}{AP}}
        & \multicolumn{1}{c|}{\multirow{2}{*}{default}}
        & \multicolumn{3}{c|}{\textbf{Resol.\,[pixel/m]}}
        & \multicolumn{3}{c|}{\textbf{Kernel size}}
        & \multicolumn{3}{c}{\textbf{Dist. step $\epsilon^{(1)}$\,[cm]}} \\
        
        &   & \text{5} & \text{10} & \text{40}
        & \text{1} & \text{5} & \text{25}
        & \text{1} & \text{20} & \text{100} \\
    \midrule
    ped.  & 23.6
         & 24.0 & 22.5 & 22.8
         & 23.4 & 23.4 & 23.4
         & 20.8 & 23.4 & 23.4 \\
    div. & 6.0
         & 3.1 & 5.1 & 6.5
         & 3.7 & 4.9 & 6.7
         & 4.8 & 4.9 & 5.0 \\
    bdry. & 26.8
         & 26.8 & 25.1 & 27.0
         & 27.8 & 27.5 & 25.9
         & 26.6 & 27.3 & 27.4 \\
    \rowcolor{gray!20}
    mean & 18.8
         & 18.0 & 17.6 & 18.8
         & 18.3 & 18.6 & 18.6
         & 17.4 & 18.5 & 18.6 \\
    \bottomrule
    \end{tabular}
    \label{tab:pipeline-abls}
\end{table}

\subsection{Parameter ablation}
\label{sec:appendix-pipeline-ablation}
For the postprocessing, there come many parameters with every filter and every algorithm we add. Thus, we mostly manually fine-tuned them based on qualitative BEV results. However, we provide an ablation for the key pipeline parameters in Tab.~\ref{tab:pipeline-abls}, using pseudo labels on the same validation set as in our main experiments. We evaluate the BEV resolution, the kernel size for the morphological dilation of the lane-markings, and  $\epsilon^{(1)}$, the initial distance threshold and step size, for polyline and polygon simplification.
A higher resolution improves the lane dividers but slightly reduces pedestrian crossing AP, which can be explained by their different shape types.
Larger dilation kernels show to significantly improve the lane preservation as fragmented lanes get connected again. We also notice that a too small $\epsilon^{(1)}$ (i.e., a too faithful approximation) harms the quality of pedestrian crossings.
To ensure full reproducibility, we published the code.

\section{Linear Program Formulation}
\label{sec:lp}
We perform both one-to-one and one-to-many assignment to optimally match elements between predictions and fragmented pseudo-labels. Thereby, we formulate a binary integer linear program with the following constraints: each pseudo-label element is assigned exactly once, and each prediction is assigned at most once. Our objective is to minimize the total matching cost.
  
Let the binary variable
\begin{equation}
x_{ij} \in \{0,1\}, \quad \forall i\in Q_\text{ind},\; j\in G_\text{ind},
\end{equation}
denotes the direct one-to-one assignment between the predicted element $q_i$ and the pseudo-label element $g_j$, and
\begin{equation}
y_{iJ} \in \{0,1\}, \quad \forall i\in Q_\text{ind},\; J \in \mathcal{J},
\end{equation}
denotes a one-to-many assignment between the predicted element $q_i$ and a set of pseudo-label elements $\{g_j\}_{j \in J}$. We enforce that every pseudo-label element should be assigned exactly once by
\begin{equation}
\sum_{i \in Q_\text{ind}} \left(x_{ij_G} + \sum_{\substack{J \in \mathcal{J} | j_G \in J}}  y_{iJ}\right) = 1, \quad \forall j_G\in G_\text{ind},
\end{equation}
and that every prediction should be assigned not more than once by 
\begin{equation}
x_i + y_i \leq 1, \quad \forall i \in Q_\text{ind}
\end{equation}
with $x_i=\sum_{j \in G_\text{ind}} x_{ij}$ as the one-to-one flag and $y_i=\sum_{J \in \mathcal{J} }  y_{iJ}$ as the one-to-many flag. The overall objective is to minimize the total cost:
\begin{equation}
\scalemath{0.8}{
\min_{\{x_{ij}\},\{y_{iJ}\}} \; \sum_{i\in Q_\text{ind}}\left( \sum_{j\in G_\text{ind}} c_{\text{o2o}}(q_i, g_j) \, x_{ij} + \sum_{J\in \mathcal{J}} c_{\text{o2m}}(S^i, J)\, y_{iJ}\right)
}
\end{equation}
yielding an optimal matching denoted as $x^*_{ij},y^*_{iJ}$.

\section{Centerlines}
\label{sec:appendix-centerlines}
\begin{figure}[t]
    \centering
    \includegraphics[width=\linewidth]{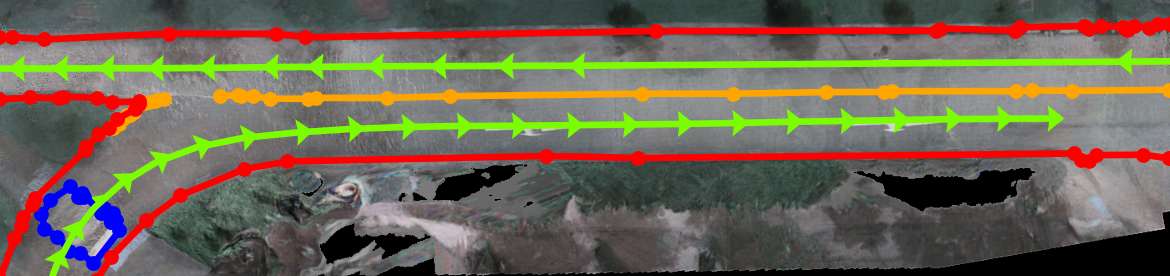}
    \caption{Centerlines derived by multi-trip ego trajectories.}
    \label{fig:centerlines}
\end{figure}

In addition to the evaluated map classes, centerlines are imaginary lines that run along the middle of driving lanes and serve as crucial references for planning. However, since we cannot infer these lines from our BEV segmentation, we suggest two potential approaches for future work: deriving centerlines from ego trajectories or extracting them from parallel polylines of existing map elements. Both methods have limitations, such as ego trajectories reflecting overtaking maneuvers and parallel polylines introducing ambiguities at intersections. A promising strategy might be combining these approaches for mutual verification. Fig.~\ref{fig:centerlines} provides a preliminary example of centerlines generated from multiple ego trajectories.

\section{Qualitative Results}
\label{sec:appendix-qualitative}
\begin{figure*}[t]
    \centering
    \begin{subfigure}[b]{0.15\linewidth}
        \centering
        \includegraphics[width=\linewidth]{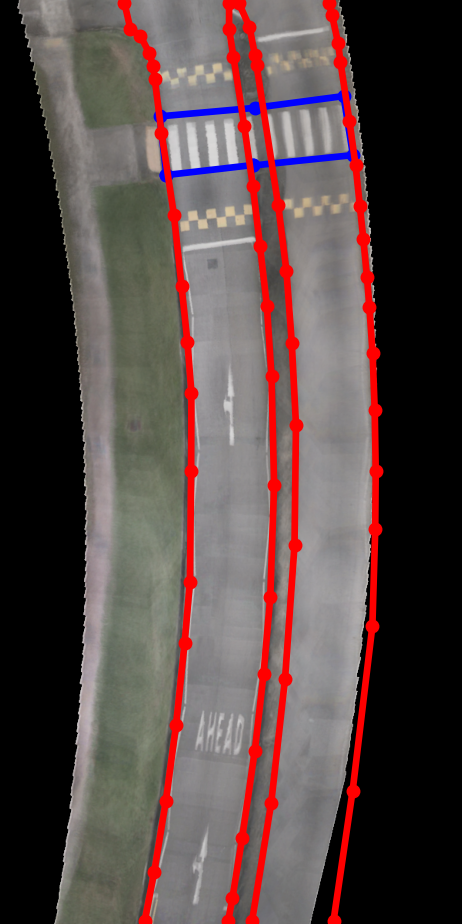}
        \caption{GT}
        \label{fig:app-1-gt}
    \end{subfigure}
    \begin{subfigure}[b]{0.15\linewidth}
        \centering
        \includegraphics[width=\linewidth]{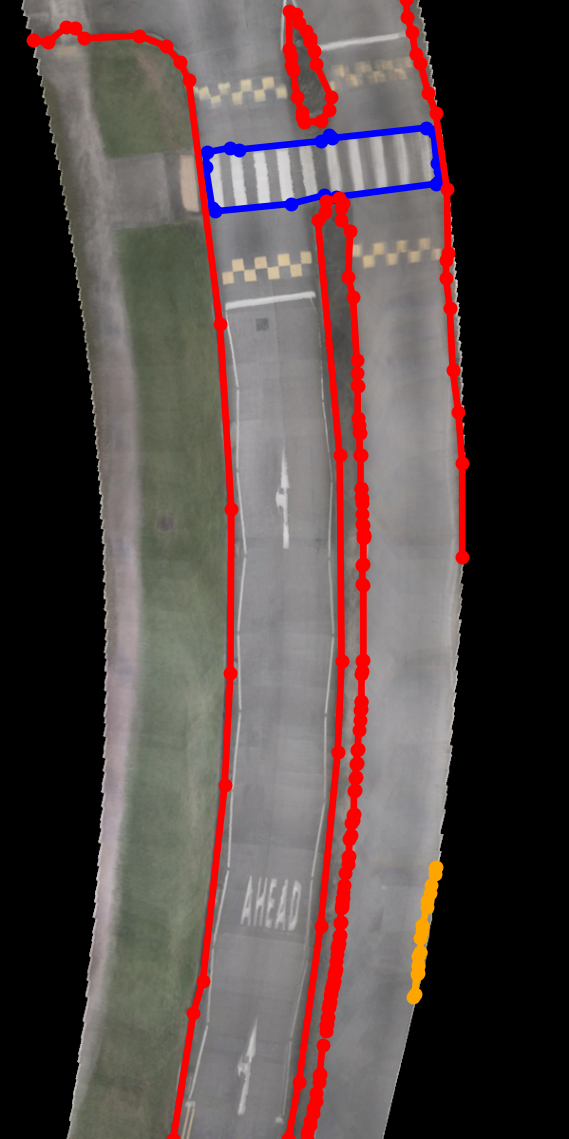}
        \caption{Single-trip}
        \label{fig:app-1-single}
    \end{subfigure}
    \begin{subfigure}[b]{0.15\linewidth}
        \centering
        \includegraphics[width=\linewidth]{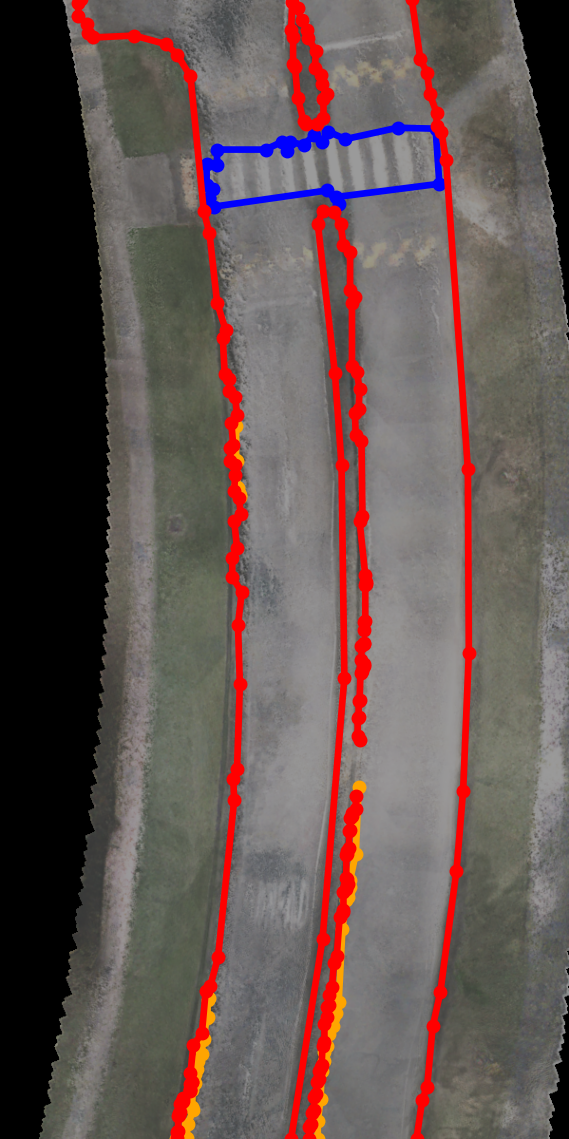}
        \caption{Multi-trip}
        \label{fig:app-1-multi}
    \end{subfigure}
    \hfill
    \begin{subfigure}[b]{0.15\linewidth}
        \centering
        \includegraphics[width=\linewidth]{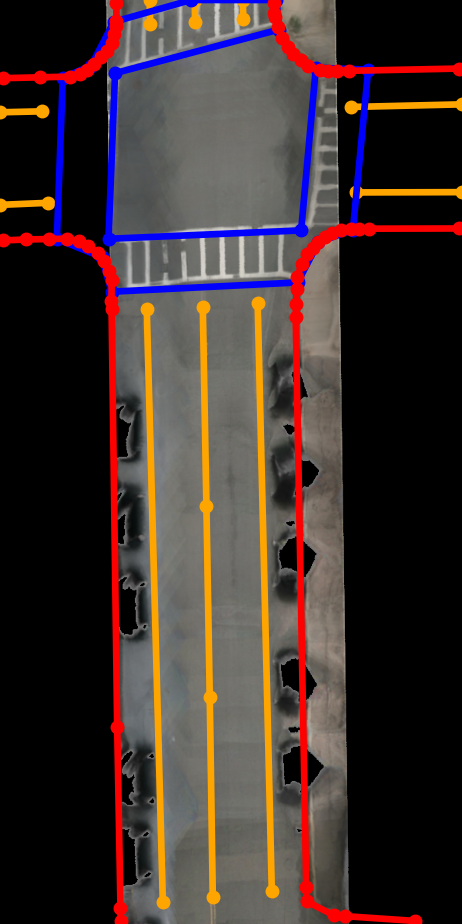}
        \caption{GT}
        \label{fig:app-3-gt}
    \end{subfigure}
    \begin{subfigure}[b]{0.15\linewidth}
        \centering
        \includegraphics[width=\linewidth]{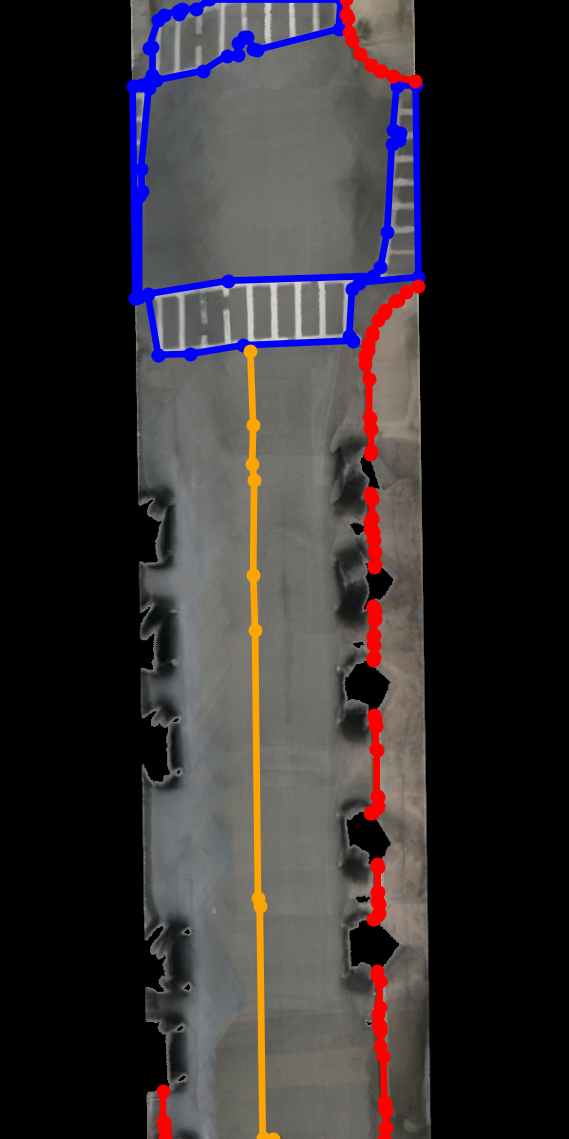}
        \caption{Single-trip}
        \label{fig:app-3-single}
    \end{subfigure}
    \begin{subfigure}[b]{0.15\linewidth}
        \centering
        \includegraphics[width=\linewidth]{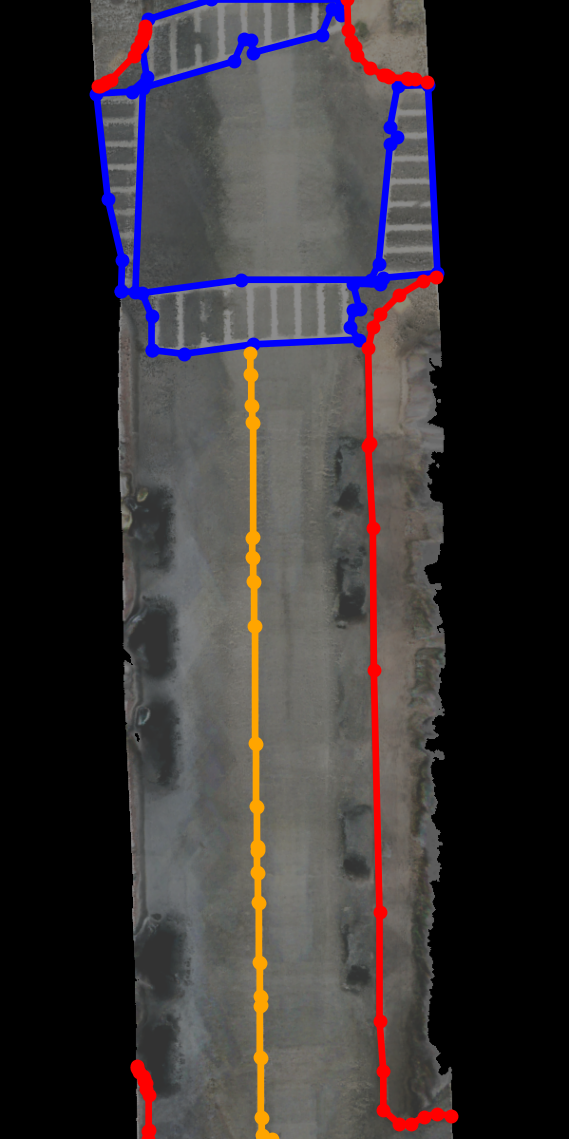}
        \caption{Multi-trip}
        \label{fig:app-3-multi}
    \end{subfigure}
    \hfill
    \begin{subfigure}[b]{0.15\linewidth}
        \centering
        \includegraphics[width=\linewidth]{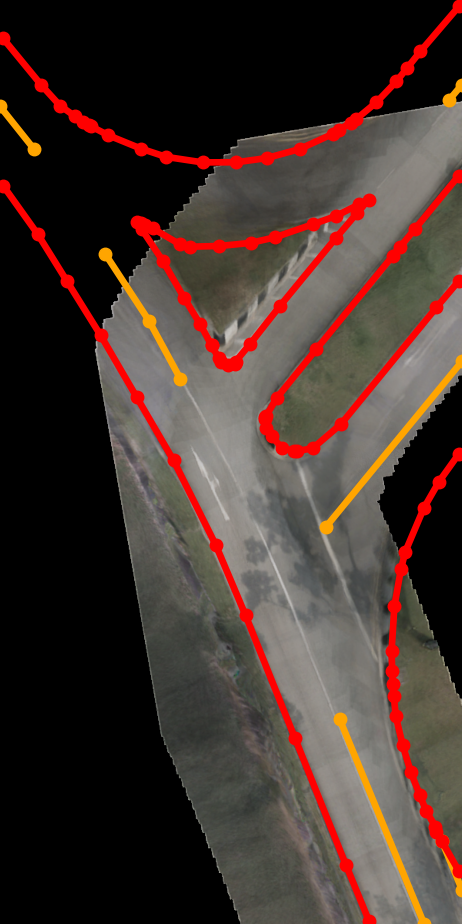}
        \caption{GT}
        \label{fig:app-2-gt}
    \end{subfigure}
    \begin{subfigure}[b]{0.15\linewidth}
        \centering
        \includegraphics[width=\linewidth]{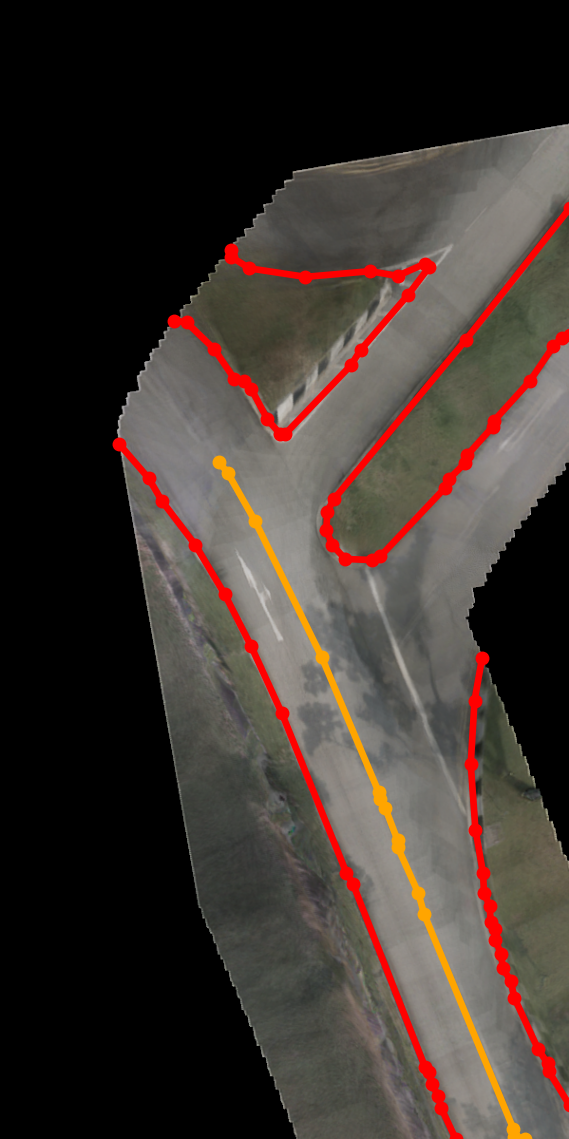}
        \caption{Single-trip}
        \label{fig:app-2-single}
    \end{subfigure}
    \begin{subfigure}[b]{0.15\linewidth}
        \centering
        \includegraphics[width=\linewidth]{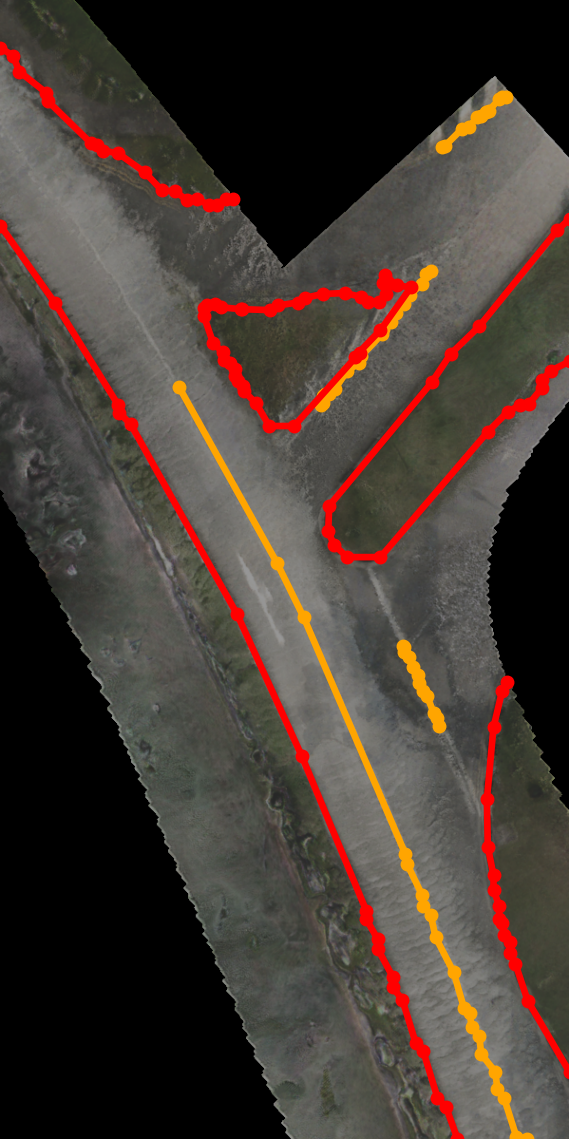}
        \caption{Multi-trip}
        \label{fig:app-2-multi}
    \end{subfigure}
    \hfill
    \begin{subfigure}[b]{0.15\linewidth}
        \centering
        \includegraphics[width=\linewidth]{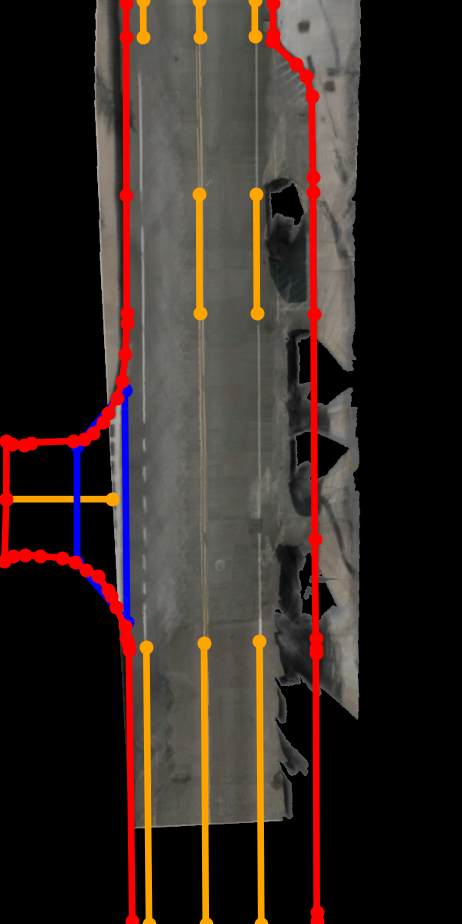}
        \caption{GT}
        \label{fig:app-4-gt}
    \end{subfigure}
    \begin{subfigure}[b]{0.15\linewidth}
        \centering
        \includegraphics[width=\linewidth]{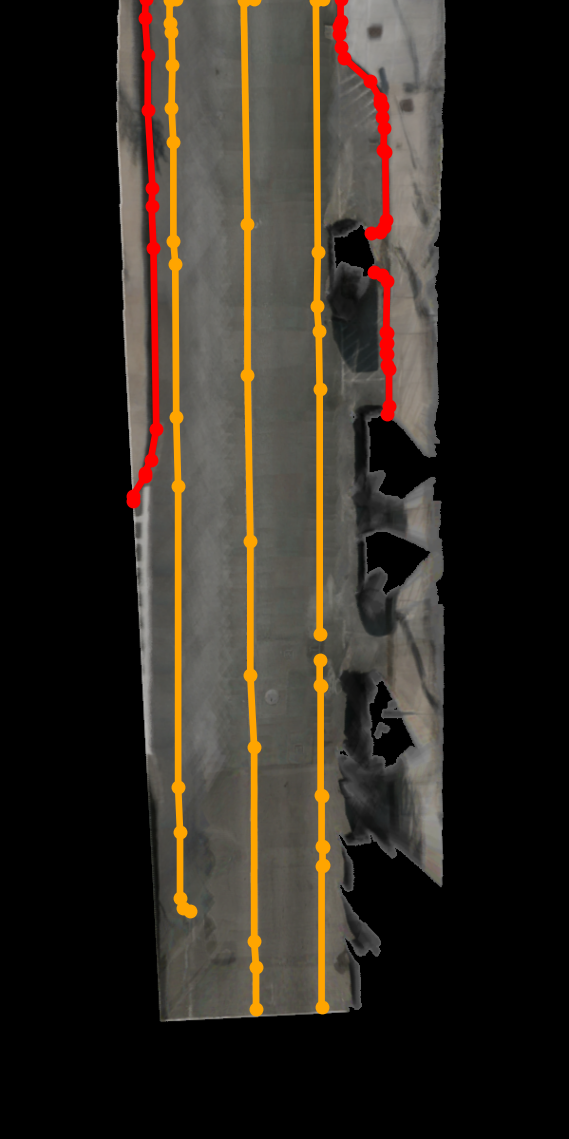}
        \caption{Single-trip}
        \label{fig:app-4-single}
    \end{subfigure}
    \begin{subfigure}[b]{0.15\linewidth}
        \centering
        \includegraphics[width=\linewidth]{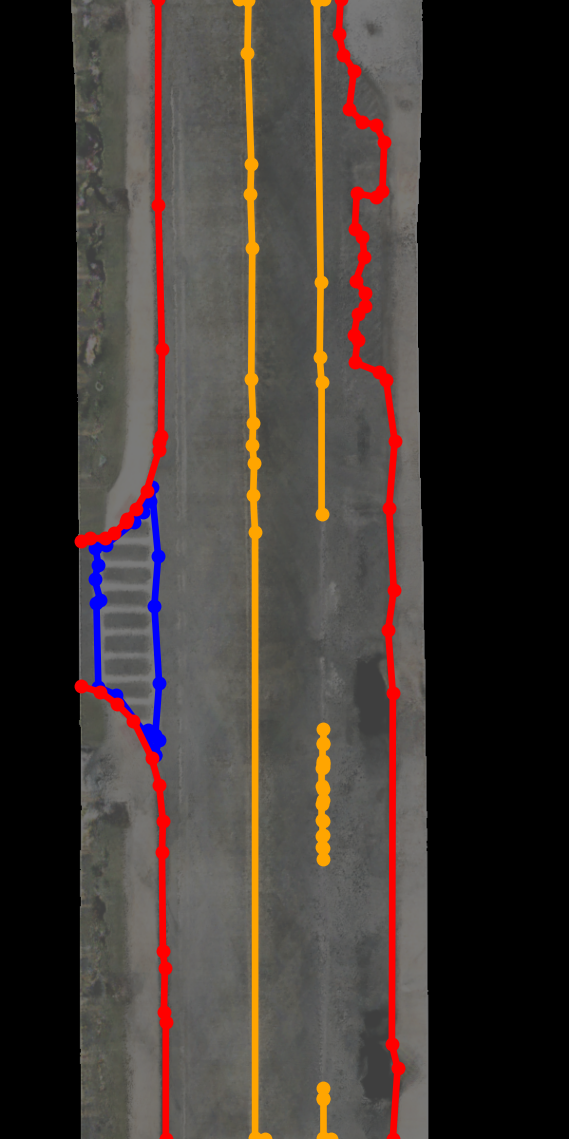}
        \caption{Multi-trip}
        \label{fig:app-4-multi}
    \end{subfigure}
    \hfill
    \begin{subfigure}[b]{0.15\linewidth}
        \centering
        \includegraphics[width=\linewidth]{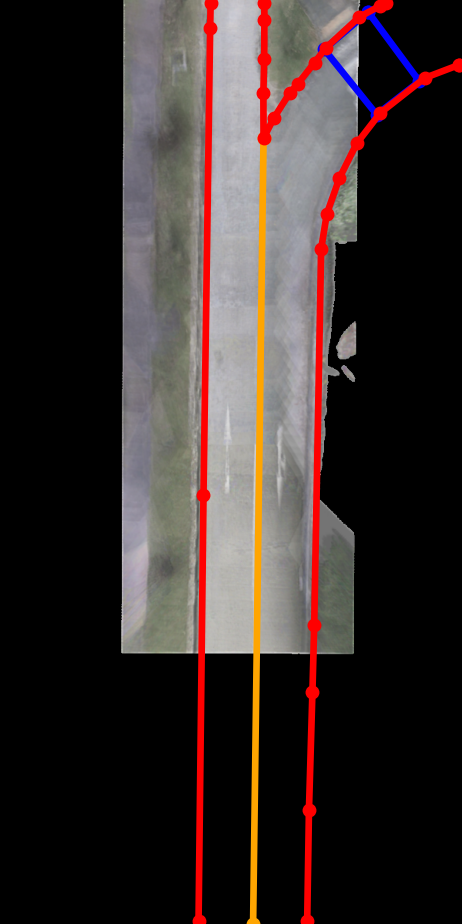}
        \caption{GT}
        \label{fig:app-6-gt}
    \end{subfigure}
    \begin{subfigure}[b]{0.15\linewidth}
        \centering
        \includegraphics[width=\linewidth]{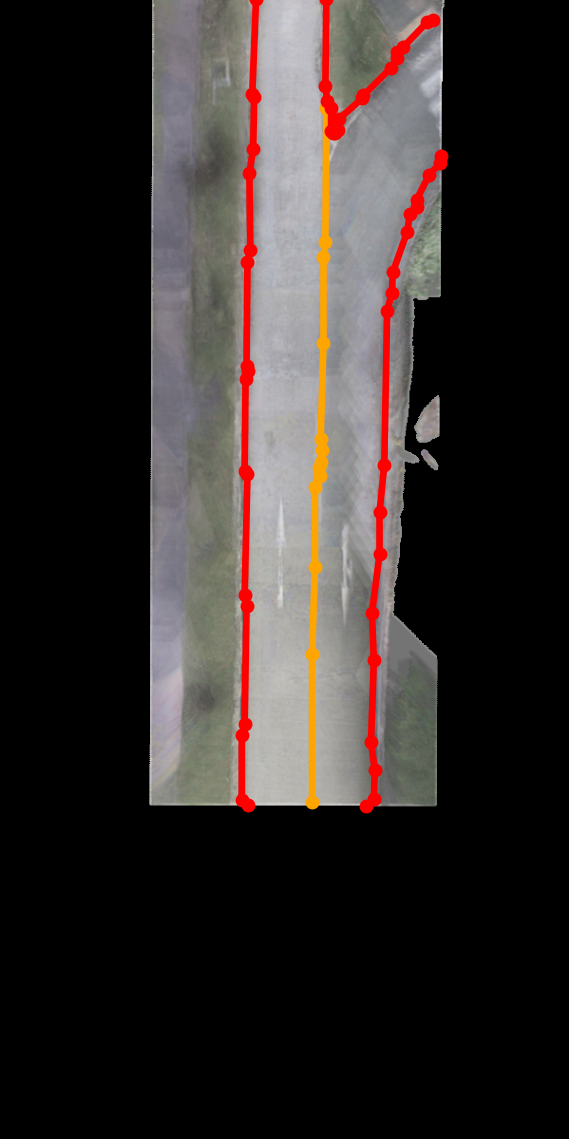}
        \caption{Single-trip}
        \label{fig:app-6-single}
    \end{subfigure}
    \begin{subfigure}[b]{0.15\linewidth}
        \centering
        \includegraphics[width=\linewidth]{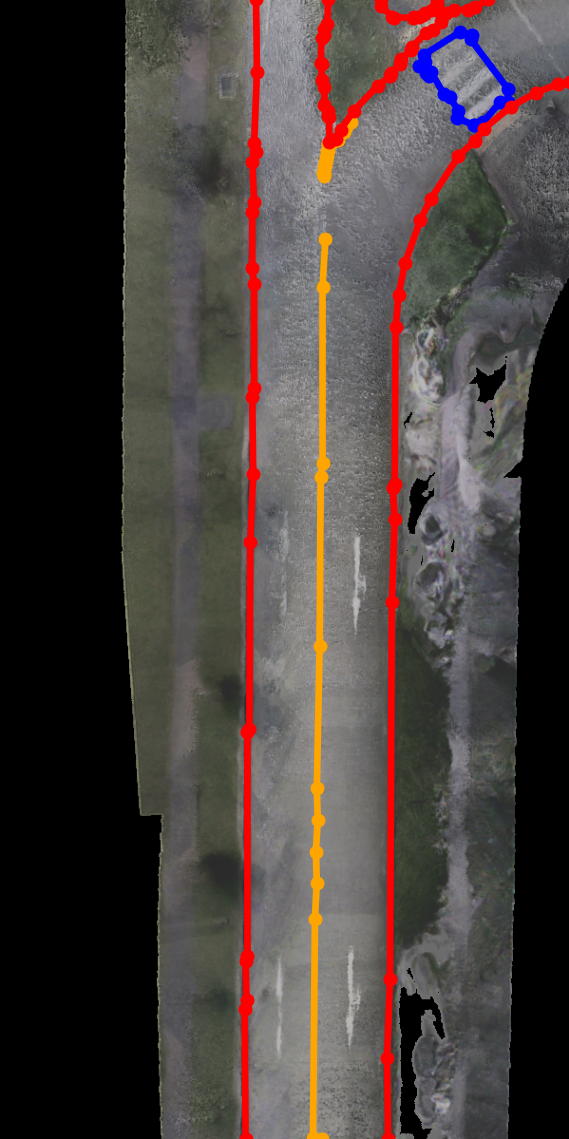}
        \caption{Multi-trip}
        \label{fig:app-6-multi}
    \end{subfigure}
    \hfill
    \begin{subfigure}[b]{0.15\linewidth}
        \centering
        \includegraphics[width=\linewidth]{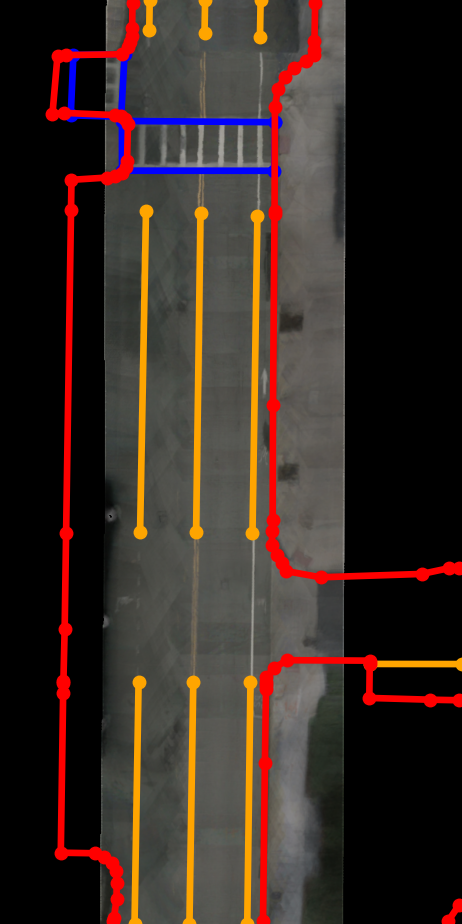}
        \caption{GT}
        \label{fig:app-5-gt}
    \end{subfigure}
    \begin{subfigure}[b]{0.15\linewidth}
        \centering
        \includegraphics[width=\linewidth]{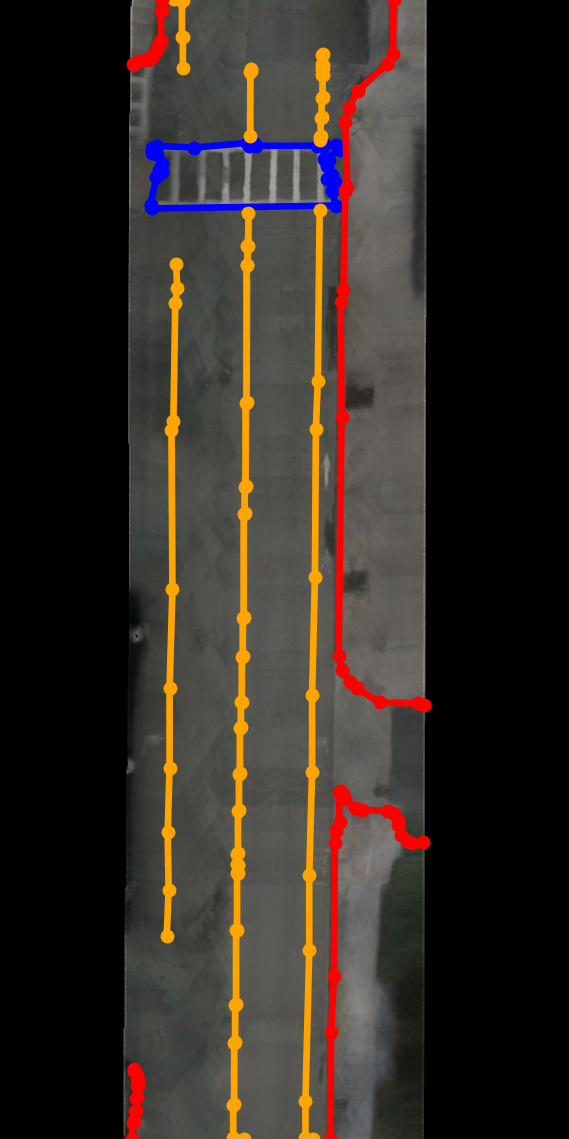}
        \caption{Single-trip}
        \label{fig:app-5-single}
    \end{subfigure}
    \begin{subfigure}[b]{0.15\linewidth}
        \centering
        \includegraphics[width=\linewidth]{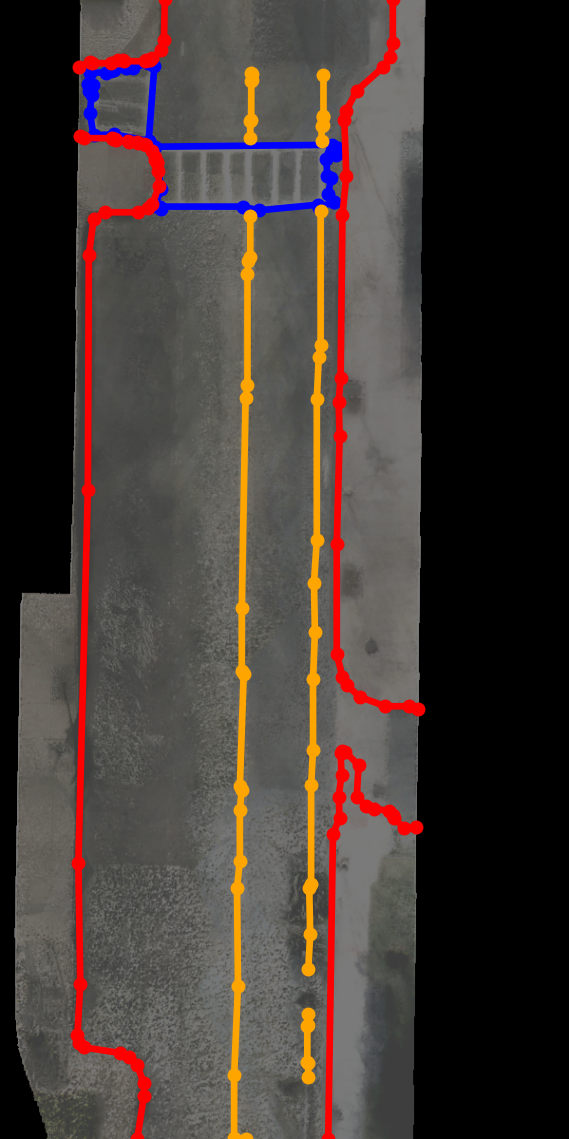}
        \caption{Multi-trip}
        \label{fig:app-5-multi}
    \end{subfigure}
    \caption{\textbf{Additional qualitative results.} We plot the lane dividers (orange), road boundaries (red), and pedestrian crossings (blue) for pseudo-labels and ground truth (GT).}
    \label{fig:app-qualitative}
\end{figure*}

Further qualitative results of our pseudo-labels compared to the ground-truth map are shown in Fig.~\ref{fig:app-qualitative}.

\section{Pseudo-Label Generation Details}
\label{sec:appendix-exp-details}
We train the Mask2Former \cite{cheng2022masked} segmentation model on 1x NVIDIA A100 GPU with a Swin-L transformer \cite{liu2021swin} with 200 queries as its backbone and pre-trained on ImageNet-21k \cite{russakovsky2015imagenet}. For the surface optimization, we also use 1x A100 GPU. For combining data from multiple trips, we limit the maximum number of trips to 50 to reduce runtime and avoid memory peaks. We group trips based on the minimum distance of their ego trajectories and deliberately exclude combinations of trips from the training set with those from the validation set. For the meshgrid expansion, we follow \cite{feng2024rogs} and choose an offset of $r=\SI{7}{m}$ along the ego trajectory.

\end{document}